\documentclass[10pt,twocolumn,letterpaper]{article}

\usepackage[pagenumbers]{wacv} 

\usepackage{arydshln} 

\usepackage[dvipsnames]{xcolor, colortbl}

\usepackage{graphicx}
\usepackage{amsmath}
\usepackage{amssymb}
\usepackage{dashrule}
\usepackage{booktabs}
\usepackage{array, multirow}
\usepackage{mathtools}
\usepackage{enumitem}
\usepackage{float}
\usepackage[normalem]{ulem}
\useunder{\uline}{\ul}{}
\graphicspath{ {./pics/} }
\usepackage{bbm}

\usepackage{pifont}
\newcommand{\cmark}{\ding{51}}
\newcommand{\xmark}{\ding{55}}

\usepackage{tabularx}
\usepackage{wrapfig}
\usepackage[bottom]{footmisc}

\usepackage{tcolorbox}
\usepackage{tikz}
\usetikzlibrary{arrows,shapes,snakes,automata,backgrounds,fit,petri}
\usepackage{adjustbox}

\definecolor{brightlavender}{rgb}{0.75, 0.58, 0.89}
\definecolor{gold}{rgb}{1.0, 0.84, 0.0}
\definecolor{skblue}{rgb}{0.0, 0.663, 1.0}
\newcommand{\gc}{\color{gray}}

\newcommand{\finalModel}{STOV-TAL}
\newcommand{\baselineModel}{STOV-TAL (w/o ST)}
\newcommand{\ovidModel}{STOV-TAL (ID-ST)}
\newcommand{\ovodModel}{STOV-TAL (OD-ST)}
\newcommand{\fullshotModel}{STOV-TAL (FS)}

\newcommand{\fVideo}{\mathbf{F}_{\mathcal{V}}}
\newcommand{\fAction}{\mathbf{F}_{\mathcal{A}}}
\newcommand{\fClass}{\mathbf{F}_{\mathcal{T}}}
\newcommand{\encVideo}{{\Phi}_{\mathcal{V}}}
\newcommand{\encText}{{\Phi}_{\mathcal{T}}}
\newcommand{\agnInst}{\mathbf{\Psi}_{agn}}
\newcommand{\clsInst}{\mathbf{\Psi}_{cls}}
\newcommand{\Inst}{\mathbf{\Psi}}

\usepackage[pagebackref,breaklinks,colorlinks]{hyperref}

\usepackage[capitalize]{cleveref}
\crefname{section}{Sec.}{Secs.}
\Crefname{section}{Section}{Sections}
\Crefname{table}{Table}{Tables}
\crefname{table}{Tab.}{Tabs.}

\def\wacvPaperID{1959} 
\def\confName{WACV}
\def\confYear{2025}

\begin{document}

\title{Exploring Scalability of Self-Training for\\ Open-Vocabulary Temporal Action Localization}

\author{
Jeongseok Hyun$^1${\qquad}Su Ho Han$^1${\qquad}Hyolim Kang$^1${\qquad} Joon-Young Lee$^2${\qquad}Seon Joo Kim$^1$
\vspace{2mm}\\$^1$Yonsei University\qquad$^2$Adobe Research
\vspace{-3mm}
}

\maketitle

\begin{abstract}
The vocabulary size in temporal action localization (TAL) is limited by the scarcity of large-scale annotated datasets. To overcome this, recent works integrate vision-language models (VLMs), such as CLIP, for open-vocabulary TAL (OV-TAL). However, despite the success of VLMs trained on extensive datasets, existing OV-TAL methods still rely on human-labeled TAL datasets of limited size to train action localizers, limiting their generalizability. In this paper, we explore the scalability of self-training with unlabeled YouTube videos for OV-TAL. Our approach consists of two stages: (1) a class-agnostic action localizer is trained on a human-labeled TAL dataset to generate pseudo-labels for unlabeled videos, and (2) the large-scale pseudo-labeled dataset is then used to train the localizer. Extensive experiments demonstrate that leveraging web-scale videos in self-training significantly enhances the generalizability of an action localizer. Additionally, we identify limitations in existing OV-TAL evaluation schemes and propose a new benchmark for thorough assessment. Finally, we showcase the TAL performance of the large multimodal model Gemini-1.5 on our new benchmark. 
Code is released at \url{https://github.com/HYUNJS/STOV-TAL}.
\end{abstract}
\section{Introduction}
\label{sec:intro}

\begin{figure}[t!]
\centering
\includegraphics[width=\linewidth]{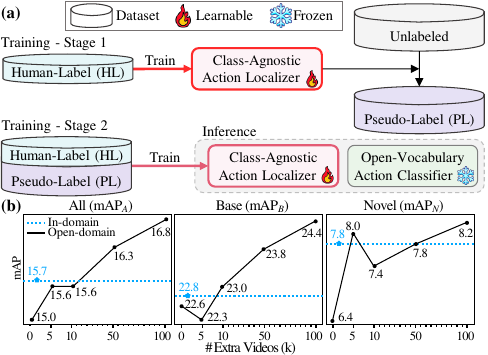}
\caption{
\textbf{(a)} Our two-stage self-training pipeline.
\textbf{(b)} Two types of data sources are explored for self-training: \textit{in-domain} (ID), including videos from novel categories of the target benchmark, and \textit{open-domain} (OD), including random web videos.
The scalability of our self-training approach is demonstrated by increasing mAPs of both base and novel actions, showing improved generalizability.
}
\label{fig:overview}
\end{figure}

Temporal action localization (TAL) involves localizing and classifying action instances in long, untrimmed videos. Labeling large TAL datasets with extensive action vocabularies is costly, limiting model vocabularies. Developing TAL models capable of generalizing to any target actions, beyond preset ones, has been a long-standing goal.

Various research directions have been explored to achieve this goal. Open-set TAL~\cite{bao2022opental} aims to localize all actions by assigning base classes and marking novel actions as unknown. Open-world TAL~\cite{zhang2023owtal} extends open-set TAL by integrating continual learning~\cite{park2021tcd}, allowing the model to be updated with annotations of novel actions after the initial training phase. However, neither of these settings is suitable for detecting new actions.

On the other hand, the zero-shot (ZS) learning approach has been integrated into TAL~\cite{zhang2020zstad, nag2023tranzad}, leveraging semantic information from language models~\cite{mikolov2013word2vec, devlin2019bert} to detect new actions.
Since the primary goal of ZS learning is to recognize concepts unseen during training~\cite{wu2024ovsurvey}, it strictly excludes any data from novel action classes during training.
This restriction hinders the improvement of TAL models' generalization ability through the use of large-scale web data.

To address the restrictive use of training data in ZS learning, Zareian \etal~\cite{zareian2021ovrcnn} introduced open-vocabulary (OV) learning for object detection, which employs image-text pairs for training. This data spans a broad language space that may overlap with test classes. In this flexible setting, vision-language models (VLMs) like CLIP~\cite{radford2021clip}, pre-trained on web image-caption pairs, are applied to various tasks, including open-vocabulary temporal action localization (OV-TAL). However, existing OV-TAL methods~\cite{ju2022effprompt, nag2022stale, phan2024zeetad, li2024detal} still rely on small, fully labeled TAL datasets for training. This approach overlooks the vast potential of web videos to enhance an action localizer. To unlock this potential, we explore self-training with unlabeled videos to improve the \textbf{generalization ability of an action localizer}. 

Fig.~\ref{fig:overview} (a) illustrates our self-training pipeline. In stage 1, a class-agnostic action localizer is trained on a human-labeled TAL dataset, generating pseudo-labels for unlabeled videos. In stage 2, the combined dataset is used to train the localizer, enhancing its generalization across diverse action categories and video domains. While class-agnostic action localizers can detect novel actions~\cite{lin2019bmn}, our self-training approach leads to improved cross-category generalizability.

We explore two distinct data sources for self-training: in-domain and open-domain. In-domain data is derived from a partition of the target benchmark, containing videos from novel categories within the same benchmark. Open-domain data consists of random web videos. While in-domain data is high-quality and closely aligned with the benchmark (often regarded as gold data), it is limited in scope. In contrast, open-domain data enables a scalable solution, as more videos can be easily scraped from the web. As shown in Fig.~\ref{fig:overview} (b), training with 100k YouTube videos outperforms in-domain data on both base and novel classes, highlighting the potential of leveraging web videos for self-training. Tab.~\ref{tab:ovtal_scalability} presents results from further scaling.

\begin{figure}[t!]
\centering
\includegraphics[width=1.0\linewidth]{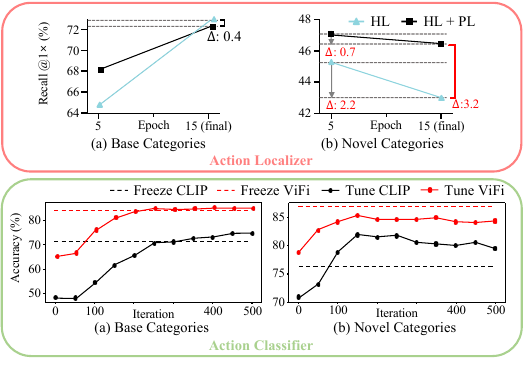}
\caption{ 
\textbf{(Top)} Training an action localizer on base categories reduces recall on novel categories. Our self-training approach mitigates this issue. \textbf{(Bottom)} Partial tuning CLIP on base actions improves accuracy on novel actions, but this improvement diminishes before base accuracy saturates.
In contrast, ViFi-CLIP, fully tuned with large video-text data, achieves better overall accuracy and does not benefit from partial tuning on a small-scale TAL dataset.
}
\label{fig:motivation}
\end{figure}

Furthermore, we study how the generalization ability changes when the action localizer and classifier are trained on a human-labeled TAL dataset of base categories. To train the action classifier, we partially fine-tune a VLM by adding a transformer layer to the visual encoder and learnable prompts to the text encoder, following EffPrompt~\cite{ju2022effprompt}. As shown in Fig.~\ref{fig:motivation}, the performance of both the localizer and classifier decreases for novel actions, while the performance on base actions saturates. Training the action localizer with our self-training approach prevents this loss of generalization ability. ViFi-CLIP~\cite{rasheed2023vificlip}, which is already fine-tuned on large video-text datasets, does not benefit from further tuning on TAL data and is, in fact, negatively affected. As a result, we use the frozen VLM directly.

The results indicate that the model with optimal performance in novel actions may be underfitted to base actions. However, current evaluation schemes~\cite{ju2022effprompt} access only randomly sampled novel actions, neglecting based actions, which can lead to potentially misleading interpretations. To ensure rigorous evaluation, we propose new OV-TAL benchmarks with two key features.

First, we introduce a generalized zero-shot setting where the target categories include both base and novel classes, and accuracies for both are reported. The base and novel classes are split according to Kinetics~\cite{carreira2017i3d}, ensuring that common actions are systematically identified rather than randomly sampled. Second, we introduce a cross-dataset evaluation to assess the generalization capability across domains. Additionally, we present results using the recent, powerful large multimodal model (LMM) Gemini 1.5 Flash/Pro~\cite{reid2024gemini}, establishing a strong baseline for LMM-based approaches in our OV-TAL benchmarks.

Our contributions can be summarized as follows:
\begin{itemize}
    \item We show the effectiveness and scalability of using unlabeled web videos for self-training to achieve cross-category and cross-domain generalization in OV-TAL.
    \item We introduce OV-TAL benchmarks for rigorous evaluation: generalized zero-shot setting with Kinetics~based category split and cross-dataset generalization.
    We present a baseline for LMM using Gemini.
\end{itemize}

\section{Related Works}
\label{sec:rel_works}

\subsection{Beyond Closed-Vocabulary TAL}

\noindent \textbf{Pre-VLM Era.}
Expanding the vocabulary of TAL models has been a longstanding goal, pursued even before the advent of VLMs. AherNet~\cite{long2020ahernet} combines TAL and action classification datasets to expand the vocabulary through adversarial learning, but its vocabulary remains limited to the human-labeled training dataset. Zhang \etal~\cite{zhang2020zstad} explored zero-shot learning in TAL to generalize to unseen actions during training by leveraging language model embeddings~\cite{mikolov2013word2vec} for both seen and unseen actions. OpenTAL~\cite{bao2022opental} tackled the open-set problem by localizing and classifying unseen actions during training as an "unknown" category, alongside seen actions. This method uses learned evidential uncertainty to determine whether a detected action should be categorized as a known class or as "unknown."

\noindent \textbf{Post-VLM Era.}
Based on VLMs~\cite{radford2021clip}, OV-TAL methods have been proposed. A decoupled framework, composed of a class-agnostic action localizer and an open-vocabulary action classifier, has been explored in~\cite{rathod2022ovtad, ju2022effprompt, ju2023multi, li2024detal}. Ju~\etal~\cite{ju2022effprompt} adapted CLIP for video by adding learnable prompts and a transformer encoder for temporal modeling.  In the follow-up~\cite{ju2023multi}, Ju \etal studied LLM-generated and visually conditioned prompts and introduced optical flow with I3D~\cite{carreira2017i3d} video backbone. Rathod~\etal~\cite{rathod2022ovtad} evaluated the performance of various image VLMs~\cite{jia2021align,radford2021clip, pham2023basic} for OV-TAL. STALE~\cite{nag2022stale} proposes a one-stage framework. UnLoc~\cite{yan2023unloc} unifies video localization tasks, including action segmentation,  moment retrieval, and TAL. ZEETAD~\cite{phan2024zeetad} adopts a dual-localization framework, utilizing both CLIP and I3D for action localization.

\noindent \textbf{Our Work.}
We explore the scalability of self-training with web videos for OV-TAL, aiming to enhance the generalization ability of the action localizer.

\subsection{Vision-Language Models}

\noindent \textbf{Image VLMs.}
CLIP~\cite{radford2021clip} shows remarkable performance in zero-shot image recognition, and the generalization ability is demonstrated across diverse domains.
This is attributed to the utilization of large-scale image-text datasets scraped from the web.
For example, CLIP~\cite{radford2021clip} and OpenCLIP~\cite{cherti2023openclip} are trained on 400M and 2B image-text pairs.
Such scalability offers benefits over manually annotated datasets, such as ImageNet-21k~\cite{deng2009imagenet} with 14M images.

\noindent \textbf{Video VLMs.}
Compared to image VLMs, video VLMs are under-explored due to a lack of large-scale video-caption datasets.
Initially, image VLMs are adapted to the video domain with prompt tuning and extra transformer layers to capture the temporal relationship~\cite{ni2022xclip, ju2022effprompt}.
ViFi-CLIP~\cite{rasheed2023vificlip} shows that simply fine-tuning CLIP in video snippets without temporal modeling improves the performance.
Open-VCLIP~\cite{weng2023openvclip} prevents the loss of generalization during video fine-tuning of CLIP through the regularization method.
MAXI~\cite{lin2023MAXI} explores the use of LLMs~\cite{li2022blip, brown2020gpt3} to generate video-caption pairs from Kinetics videos~\cite{carreira2017i3d}, while ViCLIP~\cite{wang2023internvid} is trained on InternVid~\cite{wang2023internvid}, which consists of 234M pairs and excludes videos from public datasets.

\noindent \textbf{Our Work.} We adopt ViFi-CLIP~\cite{rasheed2023vificlip} as the frozen VLM, though recently proposed VLMs can also be used.
\section{Methodology}
\label{sec:methodology}

To encompass largely available unlabeled, untrimmed web videos for training an action localizer, we adopt the open-vocabulary learning setting.
As a solution, we propose the framework, \textbf{S}elf-\textbf{T}raining for \textbf{O}pen-\textbf{V}ocabulary \textbf{T}emporal \textbf{A}ction \textbf{L}ocalization, namely, \textbf{\finalModel{}}.
This section provides a rigorous definition of this setting and the evaluation protocol.
Then, we describe the architecture of STOV-TAL and the training method that improves the generalization ability of action localization.

\begin{figure}[t!]
\centering
\includegraphics[width=1.0\linewidth]{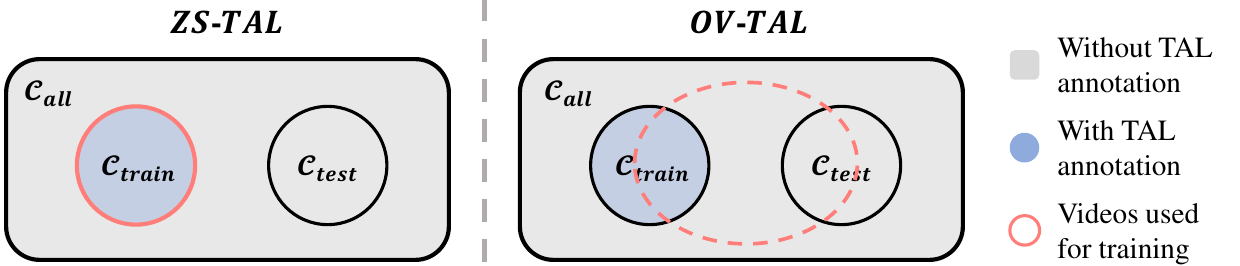}
\caption{
\textbf{Comparison of ZS-TAL and OV-TAL training.}
In ZS-TAL, the training dataset is confined to $\mathcal{C}_{train}$, which is strictly separated from $\mathcal{C}_{test}$.
OV-TAL allows the use of videos without TAL labels, even if these videos may contain actions of $\mathcal{C}_{test}$.
}
\label{fig:ovtal_setting}
\end{figure}

\subsection{Problem Definition of OV-TAL}
\label{subsec:ovtal_def}

Input consists of an untrimmed video, $\mathcal{V}$, and $C$ number of target action categories, $\mathcal{C} = \{ c_i \}_{i=1}^{C}$, which are unbounded and can be flexibly manipulated by users.
Its output consists of $M$ action instances, $\Inst = \{ (t_s, t_e, c, s)_i \}_{i=1}^{M}$, where $t_s$, $t_e$, $c$, and $s$ denote the start and end times of the action, the category, and the confidence score, respectively.

Regarding the training data utilization, the OV-TAL setting offers a more flexible approach, as depicted in Fig.~\ref{fig:ovtal_setting}.
In ZS-TAL, training data is restricted to videos with human-annotated TAL labels corresponding to $\mathcal{C}_{train}$, where $\mathcal{C}_{train} \cap \mathcal{C}_{test} = \varnothing$.
However, the OV-TAL setting relaxes this strict restriction of training data.
Specifically, we do not limit the use of data sources that may include actions of $\mathcal{C}_{test}$, as long as TAL annotations are not provided.
This enables the use of abundant unlabeled, random web videos.
As a result, training the action localizer is no longer confined to the small-scale TAL datasets, but can instead leverage data from diverse categories and domains.

\subsection{Evaluation Protocol for OV-TAL}
\label{subsec:eval_protocol}

Existing ZS-TAL benchmarks~\cite{ju2022effprompt} face two issues: (1) they do not evaluate the seen categories, but only unseen categories, and (2) they rely on random category splits that do not consider the category frequency. 
To address these, we introduce the generalized ZS evaluation setting with the Kinetics-based~\cite{carreira2017i3d} category split on the proposed OV-TAL benchmarks.
The proposed benchmarks are crucial in this blooming stage for advancing OV-TAL further.

\noindent \textbf{Generalized Zero-Shot Evaluation.}
As shown by Fig.~\ref{fig:ovtal_setting}, the benchmark dataset is disjointly divided.
In the ZS-TAL benchmark, the model is trained on the training videos of $\mathcal{C}_{train}$ and is evaluated on the validation videos of $\mathcal{C}_{test}$, \ie, $\mathcal{C}_{target} = \mathcal{C}_{test}$.
In this way, setting the target categories by either $\mathcal{C}_{train}$ or $\mathcal{C}_{test}$ is referred to the \underline{constrained} zero-shot evaluation setting.
As pointed out by Xian~\etal~\cite{xian2018zero}, the constrained setting provides strong prior knowledge and makes the problem easier by not considering $\mathcal{C}_{train}$, which are used for optimizing the model.
On the contrary, the \underline{generalized} zero-shot evaluation setting includes all categories in the benchmark for evaluation, \ie, $\mathcal{C}_{target} = \mathcal{C}_{train} \cup \mathcal{C}_{test}$.
In addition to the accuracy of all categories, we also measure the accuracy of $\mathcal{C}_{train}$ and $\mathcal{C}_{test}$ under this generalized protocol.

\noindent \textbf{Category Split.}
Existing ZS-TAL benchmarks are based on the splits whose categories are randomly chosen~\cite{zhang2020zstad, ju2022effprompt}.
Ju~\etal~\cite{ju2022effprompt} proposed ten random splits with ratios, \eg, 50\%--50\% for $\mathcal{C}_{train}$ and $\mathcal{C}_{test}$, and the final accuracy is computed by averaging the results of each split.
In our Kinetics-based split, we regard actions covered in Kinetics as common categories and use them as base categories ($\mathcal{C}_{B}$); those not covered are considered rare actions and treated as novel categories ($\mathcal{C}_{N}$).
Two reasons support this criterion:
(1) The action list in K400 is carefully selected to encompass actions that are commonly encountered in daily life. 
This selection criterion allows us to simulate a challenging scenario where categories of varying frequency - frequent for training and infrequent for testing - are utilized.
(2) K400 is a common choice for pre-training video backbones, and detecting actions not present in K400 can demonstrate a more rigorous test of category generalization ability.

To automatically detect the category overlap, we utilize NLTK~\cite{bird2009nltk} to perform stemming and lemmatization on the text of action categories and verify whether the words of action category in the benchmark are present in the set of K400 actions. 
After this process, we derive the sets of $( |\mathcal{C}_{B}|, |\mathcal{C}_{N}|)$ as $(90, 110)$, $(12, 8)$, and $(56, 50)$ for ActivityNet v1.3, THUMOS14, and FineAction, respectively.

\begin{figure*}[t!]
\centering
\includegraphics[width=1.0\linewidth]{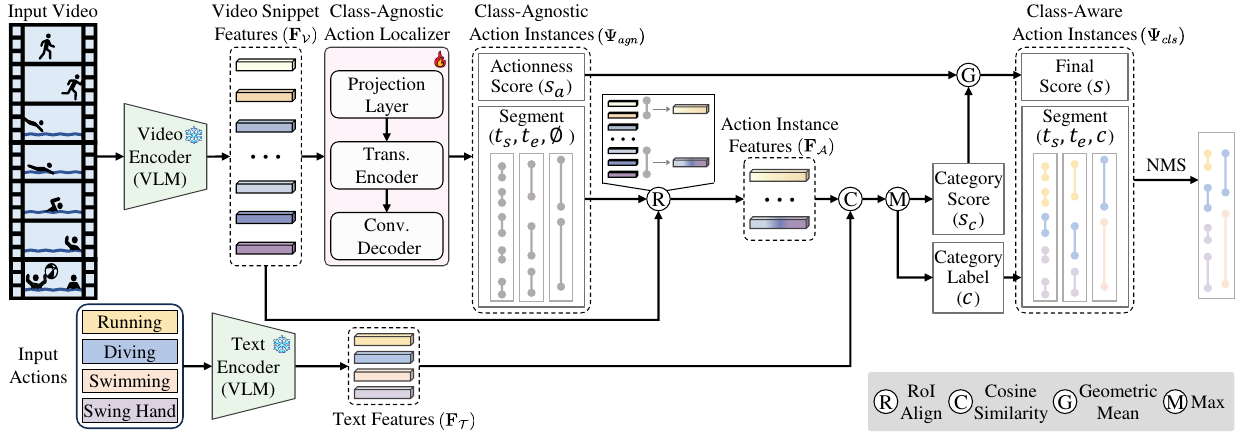}
\caption{
\textbf{Architecture.}
Features $\fVideo$ and $\fClass$ are generated by VLM video and text encoders from video frames and action names, respectively
The action localizer detects class-agnostic action instances, and their features ($\fAction$) are extracted using RoI-Align. 
Cosine similarities between $\fAction$ and $\fClass$ are computed to assign the top-scoring category to each action instance.
The category score ($s_c$) is averaged with its actionness score ($s_a$) to obtain the final confidence score ($s$).
}
\label{fig:arch}
\end{figure*}

\subsection{STOV-TAL Architecture}
\label{subsec:arch}

As illustrated in Fig.~\ref{fig:arch}, we decouple the localization and classification of actions in our architecture into two components: (1) a class-agnostic action localizer; (2) an open-vocabulary action classifier based on the video VLM~\cite{rasheed2023vificlip}. Note that this architecture supports various CLIP-like VLMs, which typically include separate uni-modal encoders, denoted as $\encVideo$ for video and $\encText$ for text.

\noindent \textbf{Video Feature Extraction with VLM.}
Frames are sampled from a video using the overlapping sliding window approach, commonly used in the TAL literature~\cite{zhang2022actionformer}. This takes consecutive frames of window size and slides a window with the stride size. As a result, a long untrimmed video is divided into multiple overlapping video snippets, represented as $\mathcal{V}=\{\mathcal{V}_i\}_{i=1}^S \in \mathbb{R}^{S \times T \times H \times W \times 3}$, where $S$ is the number of snippets and $T$ is the window size. For each snippet, the VLM's video encoder outputs the feature vector, $\encVideo (\mathcal{V}_i) \in \mathbb{R}^{D}$, where $D$ is the feature dimension. By processing the entire video, we obtain the video snippet features, $\fVideo = {\Phi}_{\mathcal{V}}(\mathcal{V}) \in \mathbb{R}^{S \times D}$.

\noindent \textbf{Detecting Every Action.}
In this stage, the goal is to detect every action instance with its actionness score $s_a$, rather than predicting its action category. To achieve this goal, we employ the action localizer that produces class-agnostic action instances, represented as $\agnInst = \{ (t_s, t_e, \varnothing, s_a)_i \}_{i=1}^{M}$. Here, $\varnothing$ indicates the absence of the action category. Our framework offers flexibility in selecting the architecture of the action localizer, provided that the output format is compatible. It accommodates both two-stage methods~\cite{lin2019bmn, xu2020gtad} and one-stage methods~\cite{lin2021afsd, zhang2022actionformer} when trained in a class-agnostic manner. The class-agnostic action localizer is improved with stronger generalization ability through the proposed self-training process, detailed in Sec.~\ref{subsec:self_training}.

To make the paper self-contained, we briefly explain the adopted action localizer~\cite{zhang2022actionformer}. This localizer consists of a convolutional projection layer, a multi-scale transformer encoder, and parallel convolutional decoders in sequence, as described in the action localizer block of Fig.~\ref{fig:arch}. The first two update the video snippet features by capturing both temporal locality and long-term contexts. Temporal downsampling is integrated into the encoder to extract multi-scale features, enabling the detecting of actions at various temporal scales due to the varying granularity of actions. Finally, two parallel convolutional decoders predict the actionness score and the offsets of start and end time for every snippet across the temporal axis at each scale.

\noindent \textbf{Classifying Every Action.}
We use an open-vocabulary action classifier for assigning the most appropriate action category ($c$) among the input actions to the previously detected action instances ($\agnInst$), leading to class-aware action instances ($\clsInst$). Following CLIP~\cite{radford2021clip}, the input actions are tokenized by the lower-cased byte pair encoding~\cite{sennrich2016bpe} with the start and end tokens. The tokenized actions are encoded through the VLM's text encoder as the text features, $\fClass \in \mathbb{R}^{C \times D}$, where $C$ is the number of input action categories. The visual features of each action instance are extracted from the VLM's video snippet features by the RoI-Align operation~\cite{he2017mask-rcnn} with its start and end times. Then, we obtain $\fAction \in \mathbb{R}^{M \times D}$, where $M$ is the number of instances.

Based on these features, we compute the cosine similarity between $\fAction$ and $\fClass$, which is a matrix of $\mathbb{R}^{M \times C}$. It is normalized by the softmax operation in the axis of the category after scaling it by the temperature parameter. This matrix value indicates each instance's class confidence scores ($s_c$). While we can choose top-$K$ action categories for each instance, we only choose the top-scoring action category as the default setting. Finally, the actionness score, $s_a$, and the category confidence score, $s_c$, of each action instance are fused by the geometric mean to compute the final confidence score, $s$. The output from this stage is class-aware action instances, represented as $\clsInst = \{ (t_s, t_e, c, s)_i \}_{i=1}^{M}$. As the post-processing, softNMS~\cite{bodla2017softnms} is applied to remove the duplicates and obtain the final action instances.

\subsection{Learning to Detect Every Action}
\label{subsec:self_training}

A class-agnostic action localizer has demonstrated the ability to localize unseen actions during training~\cite{lin2019bmn, ju2022effprompt}. Leveraging its cross-category generalization capability, our architecture can detect and classify any input actions without the need for specialized training methods. Nonetheless, its generalization ability remains limited, particularly for unseen domains (\ie, in cross-dataset evaluation). This limitation stems from the small scale of TAL datasets, which lack diversity in action categories and domains. Inspired by semi-supervised learning, a commonly approach in data-scarce scenarios, we aim to improve the action localizer's generalization ability through the self-training method~\cite{li2022rethinking, xia2023npm}. As illustrated in Fig.~\ref{fig:overview}, self-training comprises two stages: (1) Training a model on a labeled dataset; and (2) Training the model on an unlabeled dataset, using pseudo-labels generated by the previous model.

\noindent \textbf{Learning from Labeled Videos.}
In the first stage, the action localizer is trained using the human-labeled TAL dataset.
Since we employ the class-agnostic action localizer, the classification is about distinguishing the foreground or background of actions in a video.
As described in Sec~\ref{subsec:arch}, the output from the two decoders, $\agnInst$, includes the start and end times of the action and the actionness score.
These are trained with the DIoU loss~\cite{zheng2020diou} and Focal loss~\cite{lin2017focal}, respectively.

\noindent \textbf{Learning from Unlabeled Videos.}
In the second stage, we exploit additional unlabeled videos for training.
To obtain pseudo-labels of these videos, the action localizer trained in the first stage is employed to generate the class-agnostic action instances.
Note that the raw output includes many duplicated and noisy instances.
We use the NMS~\cite{bodla2017softnms} operation and the fixed threshold value on the actionness score to remove them.
Then, this pseudo-dataset is combined with the first stage's labeled dataset to form the joint dataset.
Our pseudo-label selection process requires finding the proper threshold value.
In this regard, the sensitivity analysis of pseudo-labels in Sec~\ref{subsec:ablation} supports the robustness of this process.
Although advanced pseudo-labeling methods, such as adaptive thresholding~\cite{wang2023freematch}, can be adopted, this paper focuses on \textbf{exploring the generalization ability of action localization} using the straightforward self-training method.

The process of selecting videos is crucial for self-training, where we explore two distinct settings: (1) Using \textbf{in-domain (ID) data}, and (2) Using \textbf{open-domain (OD) data}.
In the first setting, we employ the training set of the target benchmark without labels.
For instance, videos containing novel actions are chosen for self-training in the cross-category evaluation.
However, this video sampling policy is geared towards the evaluation dataset.
Utilizing videos from the same dataset ensures in-domain, creating a favorable environment for achieving higher accuracy.
Also, the scaling-up effect with a larger volume of data is constrained by a fixed and small number of videos.
On the other hand, the second setting utilizes random YouTube videos whose video IDs are sourced from K600~\cite{carreira2018k600}.
While the Kinetics dataset consists of trimmed videos focused on the specific labeled action category, we use untrimmed videos.

The training objectives used in the first stage are also used in this stage.
When the scale of the unlabeled dataset is significantly larger than that of the labeled dataset and their distributions differ, the action localizer may struggle to learn robust knowledge from the labeled dataset.
Thus, we adopt the Mean Teacher framework~\cite{tarvainen2017mean-teacher} and initialize the teacher model with the action localizer trained in the first stage.
In this teacher-student framework, the teacher model is used for inference and is updated with the student model in the manner of exponential moving average as Eq.~\ref{eq:ema}:
\begin{equation} \label{eq:ema}
    \theta_{iter}^{\prime} = (1 - \lambda) \theta_{iter} + \lambda {\theta}_{iter -1}^{\prime},
\end{equation}
where $\theta^{\prime}$ and $\theta$ represent the parameters of the teacher and student models. $\lambda$ is the smoothing coefficient, and $iter$ indicates the training iteration step.

\section{Experiments}
\label{sec:experiments}

\subsection{Datasets}
\label{subsec:datasets}

We use the following datasets for experiments.
\textbf{ActivityNet v1.3 (ANET)} ~\cite{caba2015activitynet} consists of 20,000 videos that cover 200 action categories.
Since its target categories are about activity or event, only $1$ -- $2$ of long action instances exist in a video. 
\textbf{THUMOS14 (TH14)}~\cite{idrees2017thumos14} contains 413 videos with 20 sports-related action categories.
Here, about 15 action instances exist per video on average.
\textbf{FineAction (FinAct)}~\cite{liu2022fineaction} includes 16,732 videos, encompassing 106 different action classes.
On average, each video in this dataset contains 6 action instances.
In contrast to THUMOS14, FineAction covers various domains, including not only sports but also actions related to household, socializing, and personal care activities.
Therefore, FineAction would better evaluate the generalization capability.

For self-training data, we use the video ID provided in \textbf{Kinetics-600}~\cite{carreira2018k600} and scrape full untrimmed videos from YouTube.
After filtering out very long videos exceeding 2.5 hours, about 332k videos are collected.

\subsection{Evaluation Metrics}
\label{subsec:evaluation_metrics}

We use the mean average precision (mAP), widely used to evaluate TAL methods.
For \textbf{OV-TAL benchmarks}, we report the mAP of all, base, and novel categories defined in each dataset at a temporal Intersection over Union (tIoU) of 0.5.
These metrics are denoted by $\text{mAP}^{50}_{A}$, $\text{mAP}^{50}_{B}$, and $\text{mAP}^{50}_{N}$, respectively.
For \textbf{ZS-TAL benchmarks}, we follow the conventions~\cite{ju2022effprompt, nag2022stale} reporting the mAPs at different tIoU thresholds and their average: $[0.3:0.1:0.7]$ and $[0.5:0.05:0.95]$ tIoU values for TH14~\cite{idrees2017thumos14} and ANET~\cite{caba2015activitynet}.

Across the experiments, we report the mAP of the model trained in full-shot \textbf{(FS)}, without self-training \textbf{(w/o ST)}, and with self-training (\textbf{ST}) settings.
With self-training, we present the results using in-domain and open-domain data, denoted by \textbf{(ID-ST)} and \textbf{(OD-ST)}, respectively.

\subsection{Implementation Details}
\label{subsec:impl_details}

For the video VLM, we employ ViFi-CLIP~\cite{rasheed2023vificlip} (ViFi), which is based on ViT-Base/16~\cite{dosovitskiy2021vit} visual encoder and is pre-trained on video-text pairs from the K400 dataset~\cite{carreira2017i3d}.
We use ViCLIP~\cite{wang2023internvid} trained on InternVid-10M-FLT dataset.
As pre-processing, videos are interpolated to 30 fps, resized so that the shorter side is 256 pixels, and center-cropped to 224.
To extract video features, we apply a window of 16 frames with a stride of 4 frames.
For text inputs, we directly use the action name without employing prompt engineering.

For Gemini~\cite{reid2024gemini}, we design an instruction template specifically for TAL. For video input, frames are sampled at 2 fps and resized so that the longest side is 512 pixels. To integrate fine-grained temporal information, we interleave a TimeChat-like~\cite{ren2024timechat} prompt ``The frame is sampled at \{:.2f\} seconds'' for each frame in an alternating manner.
We use Gemini 1.5 Pro/Flash (stable-001) for experimentation on OV-TAL benchmarks, which involves a cost of 1,800 USD.

\subsection{Main Results}
\label{subsec:ablation}

To assess the effectiveness of our proposed method, we perform extensive experiments on the FineAction dataset from the OV-TAL benchmark.

\begin{table}[t!]
\aboverulesep=0ex
\belowrulesep=0ex
\setlength{\tabcolsep}{3pt} 
\centering
\resizebox{1.0\linewidth}{!}{
\begin{tabular}{@{\hspace{1em}} l @{\hspace{1.0em}} | @{\hspace{1.0em}} c @{\hspace{1.0em}} c @{\hspace{1.0em}} c @{\hspace{1.0em}} | @{\hspace{1.0em}} c @{\hspace{1.0em}} c@{\hspace{1em}} }
\toprule
    \multirow{2}{*}{\begin{tabular}{@{}l@{}} Extra Videos ($\#$) \end{tabular}} &
    \multicolumn{3}{ c| @{\hspace{1.0em}}}{Generalized} & \multicolumn{2}{ c @{\hspace{1.0em}}}{Constrained}
    \\
    & mAP$_{A}^{50}$ & mAP$_{B}^{50}$ & mAP$_{N}^{50}$ & mAP$_{B}^{50}$ & mAP$_{N}^{50}$
    \\
\midrule
\midrule
    0k (w/o ST) & 15.0 & 22.6 & 6.4 & 23.3 & 7.3 \\
    \midrule
    1.5k (ID-ST) & 15.7 & 22.8 & 7.8 & 23.4 & 9.4 \\
    \cdashline{1-6} \noalign{\vskip 0.3ex}
    5k (OD-ST) & 15.6 & 22.3 & 8.0 & 23.1 & 9.7 \\
    10k (OD-ST) & 15.6 & 23.0 & 7.4 & 23.7 & 8.8 \\
    50k (OD-ST) & 16.3 & 23.8 & 7.8 & 24.6 & 9.2 \\
    100k (OD-ST) & 16.8 & 24.4 & 8.2 & 25.2 & 9.4 \\
    200k (OD-ST) & 17.0 & 24.6 & \textbf{8.5} & 25.4 & \textbf{9.9} \\
    332k (OD-ST) & \textbf{17.2} & \textbf{25.1} & 8.4 & \textbf{25.8} & 9.5 \\
    \midrule
    \gc Full-Shot (FS) & \gc 18.8 & \gc 23.5 & \gc 13.5 & \gc 24.5 & \gc 16.0 \\
\bottomrule    
\end{tabular}
}
\caption{
\textbf{Scalability of self-training with varying sizes of videos.}
}
\label{tab:ovtal_scalability}
\end{table}

\noindent \textbf{Scalability Evaluation.} 
Tab.~\ref{tab:ovtal_scalability} shows that self-training using open-domain data achieves increasing mAP$_{B}^{50}$ and mAP$_{N}^{50}$ with larger volume and outperforms the results of using in-domain data.
We achieve these noteworthy results without a specially designed loss function or pseudo-labeling strategy.
This demonstrates the high potential of self-training for OV-TAL with large-scale web videos.

The categories of scraped videos most likely belong to $\mathcal{C}_{B}$ since Kinetics dataset~\cite{carreira2017i3d} is used for sourcing videos for self-training and deciding $\mathcal{C}_{B}$.
Thus, the improvement in $\text{mAP}_{B}$ is consistent, compared to fluctuating $\text{mAP}_{N}$.

\begin{table}[t!]
\aboverulesep=0ex
\belowrulesep=0ex
\setlength{\tabcolsep}{3pt} 
\centering
\resizebox{0.95\linewidth}{!}{
\begin{tabular}{l|c|ccc|cc}
\toprule
    \multirow{2}{*}{VLM} & \multirow{2}{*}{\shortstack{Video-text \\ Pre-training}} &
    \multicolumn{3}{c|}{Generalized} & \multicolumn{2}{c}{Constrained}
    \\
    & & $\text{mAP}^{50}_{A}$ & $\text{mAP}^{50}_{B}$ & $\text{mAP}^{50}_{N}$ & $\text{mAP}^{50}_{B}$ & $\text{mAP}^{50}_{N}$ 
    \\
\midrule
\midrule
    CLIP-B~\cite{radford2021clip} & \xmark & 10.3 & 15.2 & 4.8 & 16.0 & 5.7 \\
    ViCLIP-B~\cite{wang2023internvid} & \cmark & 11.6 & 16.2 & \textbf{6.5} & 17.6 & 7.6 \\
    ViCLIP-L~\cite{wang2023internvid} & \cmark & 12.9 & 18.7 & \textbf{6.5} & 19.9 & \textbf{7.9} \\
    ViFi-B~\cite{rasheed2023vificlip} & \cmark & \textbf{15.0} & \textbf{22.6} & 6.4 & \textbf{23.3} & 7.3 \\
\bottomrule
\end{tabular}
}
\caption{\textbf{Ablation study of VLMs as backbone.}}
\label{tab:abl_vlm}
\end{table}
\begin{table}[t!]
\aboverulesep=0ex
\belowrulesep=0ex
\setlength{\tabcolsep}{3pt}
\centering
\resizebox{0.8\linewidth}{!}{
\begin{tabular}{l|ccc|cc}
\toprule
    \multirow{2}{*}{Scoring Method} &
    \multicolumn{3}{c|}{Generalized} & \multicolumn{2}{c}{Constrained}
    \\
    & $\text{mAP}^{50}_{A}$ & $\text{mAP}^{50}_{B}$ & $\text{mAP}^{50}_{N}$ & $\text{mAP}^{50}_{B}$ & $\text{mAP}^{50}_{N}$ 
    \\
\midrule
\midrule
    Category Score ($s_c$) & 1.0 & 1.4 & 0.5 & 1.3 & 0.5 \\
    Actionness Score ($s_a$) & 13.9 & 21.7 & 5.2 & 22.4 & 6.0 \\
    Arithmetic Mean & 13.5 & 20.5 & 5.6 & 21.4 & 6.7 \\
    Geometric Mean & \textbf{15.0} & \textbf{22.6} & \textbf{6.4} & \textbf{23.3} & \textbf{7.3} \\
\bottomrule
\end{tabular}
}
\caption{\textbf{Ablation study of score fusion methods.}}
\label{tab:abl_score_fusion}
\end{table}

\noindent \textbf{VLM Backbone.}
In our decoupled architecture, different VLMs can be flexibly adopted for STOV-TAL. As shown in Tab.~\ref{tab:abl_vlm}, both ViCLIP and ViFi outperform CLIP, benefiting from large-scale video-text data. ViCLIP shows high mAP$_{N}$, while ViFi exhibits high mAP$_{B}$ since it is trained on Kinetics data, whereas ViCLIP is trained on newly crawled YouTube videos. We expect that future VLMs trained on even larger dataset will further boost performance, underscoring the expandibility of our architecture.

\noindent \textbf{Score Fusion.}
Tab.~\ref{tab:abl_score_fusion} studies the score fusion method used in the inference time.
The mAP significantly decreases without relying on the actionness score since the background category is not modeled in our frozen VLM.
The arithmetic mean operation shows the lower mAPs than only using the $s_a$ since some high $s_c$ may dominate the final score.
The different scale between $s_a$ and $s_c$ is caused by the different normalization operations, \eg,~sigmoid and softmax functions.
On the other hand, the geometric mean operation alleviates this problem and shows the best mAP.

\begin{figure}[t!]
\centering
\includegraphics[width=0.9\linewidth]{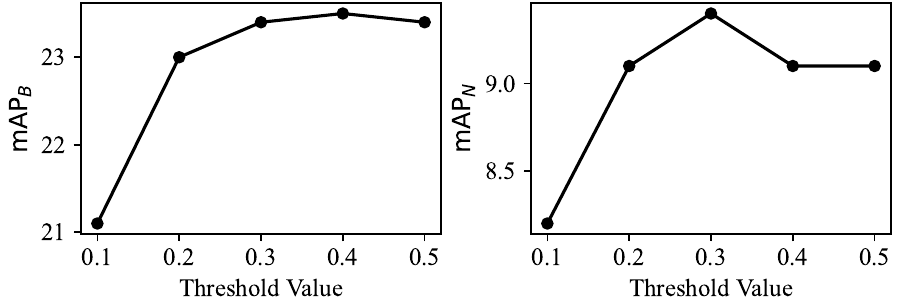}
\vspace{-2mm}
\caption{
\textbf{Sensitivity analysis of pseudo-labels in self-training.}
}
\vspace{-3mm}
\label{fig:pl_sensitivity}
\end{figure}
\begin{figure}[t!]
\centering
\includegraphics[width=1.0\linewidth]{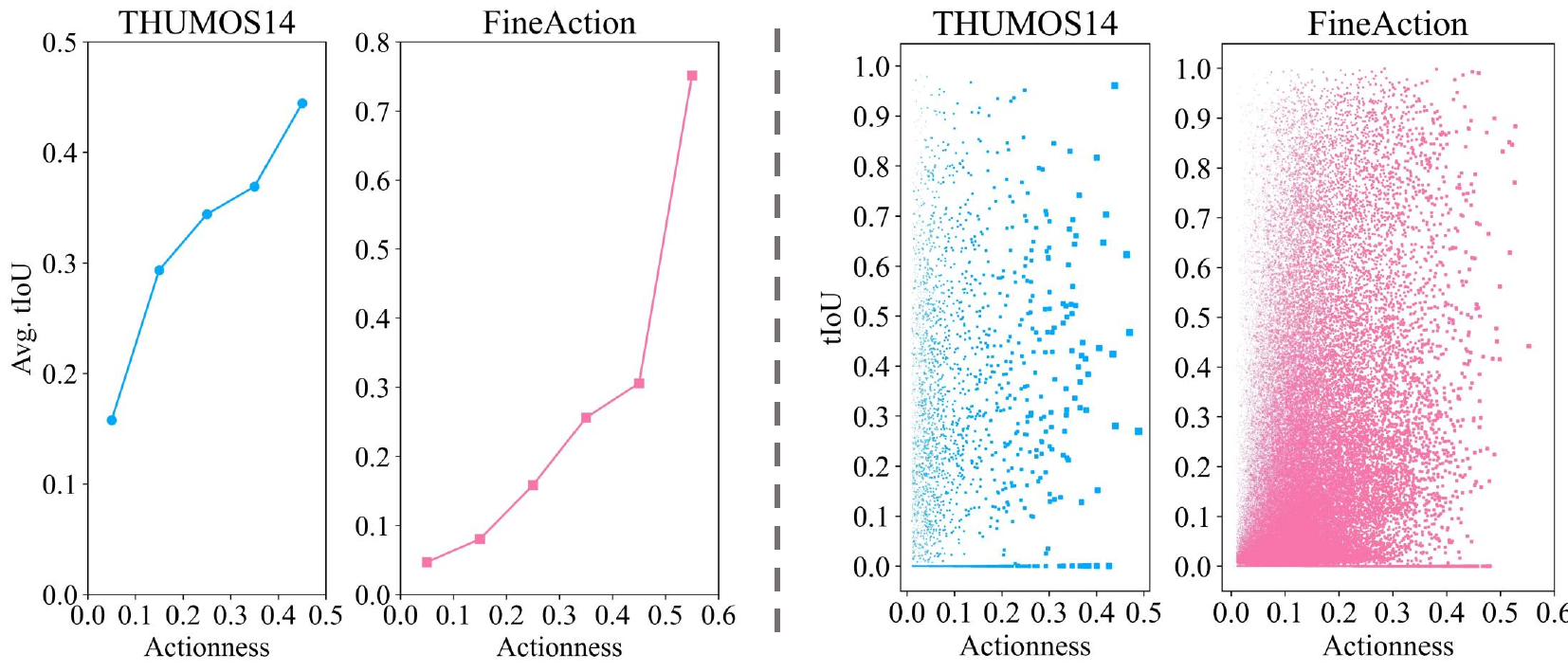}
\vspace{-4mm}
\caption{
\textbf{Quality analysis of pseudo-labels.}
We measure the correlation between the actionness of pseudo-labels and their tIoU with GT instances.
The left shows the average tIoU values for each actionness interval, while the right presents a scatter plot
}
\vspace{-3mm}
\label{fig:pl_quality}
\end{figure}

\noindent \textbf{Pseudo-labeling for Self-training.}
We apply a threshold during pseudo-labeling to filter out low-actionness action instances, thereby improving the reliability of the pseudo-labels. As shown in Fig.~\ref{fig:pl_sensitivity}, once the threshold value surpasses a certain point, this intuitive pseudo-labeling method maintains consistent performance across different threshold values. Furthermore, Fig.~\ref{fig:pl_quality} illustrates a positive correlation between the actionness of pseudo-labels and their tIoU with ground truth instances. It also reveals that low-actionness predictions tend to cluster at low tIoU values.
 
While advanced self-training techniques for TAL~\cite{xia2023npm, zhou2024apl} could be explored, this paper focuses on evaluating the scalability of the self-training action localizer and demonstrating the potential of web videos to expand the dataset's diversity and scale for OV-TAL, rather than on developing a novel pseudo-labeling method.

\subsection{Results of OV-TAL Benchmark}

Due to the absence of reproducible public code of previous OV-TAL methods~\cite{nag2022stale, ju2022effprompt}, we report the results based on our reproduced implementation.
For OpenTAL~\cite{bao2022opental}, we use our VLM-based action classifier on top of its localizer with I3D backbone.
For EffPrompt~\cite{ju2022effprompt}, we use our class-agnostic action localizer with the prompt-tuned VLM (\eg,ViFi) as an action classifier.
The details are in the supp.

\begin{table*}[t!]
\aboverulesep=0ex
\belowrulesep=0ex
\centering
\resizebox{0.95\linewidth}{!}{
\begin{tabular}{l|l|l| ccc | cc | ccc | cc }
    \toprule
    \multicolumn{2}{c|}{\multirow{3}{*}{Methods}} & \multirow{3}{*}{Backbone} & \multicolumn{5}{c|}{THUMOS14} & \multicolumn{5}{c}{FineAction}
    \\
    \cmidrule{4-8} \cmidrule{9-13}
    \multicolumn{2}{c|}{} & & \multicolumn{3}{c|}{Generalized} & \multicolumn{2}{c|}{Constrained} &
    \multicolumn{3}{c|}{Generalized} & \multicolumn{2}{c}{Constrained}
    \\
    \multicolumn{2}{c|}{} & & mAP$_{A}^{50}$ & mAP$_{B}^{50}$ & mAP$_{N}^{50}$ & mAP$_{B}^{50}$ & mAP$_{N}^{50}$ & mAP$_{A}^{50}$ & mAP$_{B}^{50}$ & mAP$_{N}^{50}$ & mAP$_{B}^{50}$ & mAP$_{N}^{50}$ 
    \\
\midrule
\midrule
\multirow{6}{*}{\begin{tabular}[c]{@{}l@{}}Non-LMM\end{tabular}}
&    OpenTAL$^\dagger$~\cite{bao2022opental} & I3D~\cite{carreira2017i3d}  & 30.2 & 42.7 & 11.3 & 42.6 & 11.9 &
            6.2 & 10.4 & 1.4 & 11.0 & 1.8
            \\
&    EffPrompt$^\dagger$~\cite{ju2022effprompt} & ViFi-B~\cite{rasheed2023vificlip} & 41.3 & 62.6 & 9.3 & 64.0 & 16.5 & 13.3 & 22.5 & 3.0 & 23.8 & 5.5
    \\
&    \textbf{\baselineModel{}} & ViFi-B~\cite{rasheed2023vificlip}  & 45.7 & \textbf{66.0} & 15.3 & \textbf{65.9} & 16.2 & 
            15.0 & 22.6 & 6.4 & 23.3 & 7.3
            \\
&    \textbf{\ovidModel{}} & ViFi-B~\cite{rasheed2023vificlip} & \textbf{46.6} & 65.8 & \textbf{17.9} & 65.8 & \textbf{18.2} &
            15.7 & 22.8 & 7.8 & 23.4 & 9.4
            \\ 
&    \textbf{\ovodModel{}} & ViFi-B~\cite{rasheed2023vificlip} &
44.2 & 62.6 & 16.7 & 62.5 & 17.0
&            \textbf{17.2} & \textbf{25.1} & \textbf{8.4} & \textbf{25.8} & \textbf{9.5}
            \\
\cmidrule{2-13}
&    \gc \fullshotModel{} & \gc ViFi-B~\cite{rasheed2023vificlip}  & \gc 50.2 & \gc 65.8 & \gc 26.8 & \gc 66.0 & \gc 27.5 & \gc 18.8 & \gc 23.5 & \gc 13.5 & \gc 24.5 & \gc 16.0 \\
\midrule
\multirow{2}{*}{\begin{tabular}[c]{@{}l@{}}LMM\end{tabular}}
&     Gemini 1.5 Flash~\cite{reid2024gemini} &  Unknown &  10.4 &  15.3 &  3.0 &  13.5 &  3.0 &  10.0 &  7.5 &  12.7 &  8.7 &  13.6
    \\
&     Gemini 1.5 Pro~\cite{reid2024gemini} &  Unknown &  19.9 &  26.4 &  10.2 &  27.4 &  13.9 &  13.1 &  11.1 &  15.3 &  11.3 &  16.6
    \\
\bottomrule
\end{tabular}
}
\caption{
\textbf{Evaluation of cross-category OV-TAL benchmark}.
For w/o ST, the model is trained on $\mathcal{C}_{B}$ of target benchmark (TH14/FinAct).
For ID-ST, training videos of $\mathcal{C}_{N}$ of the target benchmark are used.
For OD-ST, web-scraped videos are utilized.
Full-shot trained results are shown in \textcolor{gray}{gray}.
$^\dagger$ is our reproduced result.
For Gemini, some videos are blocked and excluded from evaluation.
}
\label{tab:ovtal_cross_cateogry}
\vspace{-3mm}
\end{table*}

\begin{table}[t!]
\aboverulesep=0ex
\belowrulesep=0ex
\centering
\resizebox{1.0\linewidth}{!}{
\begin{tabular}{l|l| ccc | cc }
    \toprule
    \multirow{2}{*}{Methods} & \multirow{2}{*}{Backbone} &
    \multicolumn{3}{c|}{Generalized} & \multicolumn{2}{c}{Constrained} \\
    & & mAP$_{A}^{50}$ & mAP$_{B}^{50}$ & mAP$_{N}^{50}$ & mAP$_{B}^{50}$ & mAP$_{N}^{50}$ 
    \\
\midrule
\midrule
    EffPrompt$^\dagger$~\cite{ju2022effprompt} & ViFi-B~\cite{rasheed2023vificlip} & 26.2 & 38.4 & 8.0 & 38.6 & 9.0
    \\
    \textbf{\baselineModel{}} & ViFi-B~\cite{rasheed2023vificlip} & 29.7 & 42.1 & 11.1 & 42.1 & 11.4
    \\ 
    \textbf{\ovidModel{}} & ViFi-B~\cite{rasheed2023vificlip} & 30.9 & 43.6 & \textbf{11.9} & 43.5 & 11.6
    \\
    \textbf{\ovodModel{}} & ViFi-B~\cite{rasheed2023vificlip} & \textbf{31.8} & \textbf{45.4} & 11.5 & \textbf{45.5} & \textbf{12.0}
    \\
\bottomrule
\end{tabular}
}
\caption{
\textbf{Evaluation of cross-dataset OV-TAL benchmark}.
The models trained using FinAct or YouTube videos are evaluated on TH14.
$^\dagger$ is our reproduced result.
}
\label{tab:ovtal_cross_dataset}
\vspace{-2mm}
\end{table}

\noindent \textbf{Cross-category Evaluation.}
As shown in Tab.~\ref{tab:ovtal_cross_cateogry}, the results of EffPrompt highlight the limitation of the constrained setting, which evaluates only $\mathcal{C}_{B}$ or $\mathcal{C}_{N}$ separately. Specifically, in TH14, the mAP$_{N}$ is 16.5, while in the generalized setting, it drops significantly to 9.3. In contrast, others exhibit only a small drop. Since both EffPrompt and STOV-TAL (w/o ST) use the same action localizer, this discrepancy is likely caused by tuning the VLM on $\mathcal{C}_{B}$, which biases the model toward $\mathcal{C}_{B}$ when both $\mathcal{C}_{B}$ and $\mathcal{C}_{N}$ are present for testing. This loss of generalizability is not revealed in the existing benchmark, which only evaluates $\mathcal{C}_{N}$ under the constrained setting. Therefore, our OV-TAL benchmark is crucial for thorough evaluation of generalizability.

Self-training with either ID or OD data leads to improvement in mAP$_{N}^{50}$. On TH14, mAP$_{N}^{50}$ increases from 15.3 to 17.9 with ID data, and to 16.7 with OD data. On FinAct, it rises from 6.4 to 7.8 and 8.4, respectively. These results demonstrate that our self-training method enhances the action localizer’s cross-category generalization. 

While self-training with OD data leads to higher mAP$_{B}^{50}$ in FinAct, it results in a decrease in TH14. This discrepancy stems from the differing nature of the datasets: FinAct contains videos from diverse domains covering various actions, whereas TH14 is limited to sports videos. Training with OD data introduces broader generalization capabilities, as shown by improved mAP$_{B}^{50}$ and mAP$_{N}^{50}$ in FinAct. However, in TH14, which evaluates a narrow range of actions, this broader generalization may reduce the model's specialization in $\mathcal{C}_{B}$ of TH14. 

We also experiment with the recent powerful proprietary LMM Gemini 1.5 Flash and Pro. While Gemini substantially underperforms in TH14, it outperforms our full-shot trained model for $\mathcal{C}_{N}$ in FinAct.
We speculate these are related to the different duration of action instances. 
The average duration of action instances of TH14's ($\mathcal{C}_{B}$, $\mathcal{C}_{N}$) and FinAct's ($\mathcal{C}_{B}$, $\mathcal{C}_{N}$) are (5.1, 2.6) and (7.5, 12.4) seconds, respectively. As FinAct's action instances of $\mathcal{C}_{N}$ are generally longer than others, Gemini performs well on this subset. In contrast, Gemini underperforms on TH14's $\mathcal{C}_{N}$ since action instances are too short for Gemini to detect precisely, considering that Gemini takes frames every 0.5 seconds. Detecting short action instances using LMMs could be promising future work. 

\noindent \textbf{Cross-dataset Evaluation.}
Tab.~\ref{tab:ovtal_cross_dataset} presents the results of cross-dataset evaluation. Under both ID-ST and OD-ST settings, we can observe the improvement of both $\mathcal{C}_{B}$ and $\mathcal{C}_{N}$, compared to the results without self-training. mAP$_{A}^{50}$ of OD-ST is higher than that of ID-ST, indicating better cross-domain generalization through self-training with OD data. These results demonstrate the enhanced cross-domain generalization ability from the self-training with web videos.

\begin{table}[t!]
\aboverulesep=0ex
\belowrulesep=0ex
\setlength{\tabcolsep}{3pt} 
\centering
\resizebox{1.0\linewidth}{!}{
\begin{tabular}{ l|l|l| ccc|c | cc|c }
\toprule
  \multirow{2}{*}{\begin{tabular}{@{}l@{}} Setting \end{tabular}} &
  \multirow{2}{*}{Methods} &
  \multirow{2}{*}{Backbone} &
  \multicolumn{4}{c|}{THUMOS14} & 
  \multicolumn{3}{c}{ActivityNet} \\
  \cmidrule{4-7} \cmidrule{8-10} 
  & & & 0.3 & 0.5 & 0.7 & Avg. & 0.5 & 0.75 & Avg. \\
\midrule
\midrule
\multirow{9}{*}{\begin{tabular}[c]{@{}l@{}}75\%\\Seen\\25\%\\Unseen \end{tabular}}
    & EffPrompt~\cite{ju2022effprompt} & CLIP-B~\cite{radford2021clip} & 
    39.7 & 23.0 & 7.5 & 23.3 & 
    37.6 & 22.9 & 23.1 \\
    & STALE~\cite{nag2022stale} & CLIP-B~\cite{radford2021clip} & 
    40.5 & 23.5 & 7.6 & 23.8 &
    38.2 & 25.2 & 24.9 \\
    & UnLoc~\cite{yan2023unloc} & CLIP-B~\cite{radford2021clip} &
    - & - & - & - &
    40.2 & - & - \\
    & UnLoc~\cite{yan2023unloc} & CLIP-L~\cite{radford2021clip} &
    - & - & - & - &
    48.8 & - & -\\
    & Ju~\etal~\cite{ju2023multi} & CLIP-B~\cite{radford2021clip} &
    46.3 & 29.5 & 8.7 & 28.4 &
    42.0 & 25.8 & 25.9 \\
    \cmidrule(lr){2-10}
    & \textbf{\baselineModel{}} & CLIP-B~\cite{radford2021clip} & 
    47.8 & 28.4 & 9.1 & 28.4 & 
    47.0 & 28.1 & 27.9 \\   
    & \textbf{\baselineModel{}} & ViFi-B~\cite{rasheed2023vificlip} & 
    56.7 & 34.3 & 11.3 & 34.5 &
    51.7 & \textbf{30.9} & \textbf{30.5} \\
    & \textbf{\ovidModel{}} & ViFi-B~\cite{rasheed2023vificlip} & 
    \textbf{59.5} & \textbf{37.5} & \textbf{12.5} & \textbf{36.9} &
    \textbf{52.0} & 30.6 & 30.1 
    \\ 
    & \textbf{\ovodModel{}} & ViFi-B~\cite{rasheed2023vificlip} &
    58.2 & 35.1 & 11.8 & 35.2 &
    - & - & - \\
    \cmidrule(l){2-10}
    & \gc \fullshotModel{} & \gc ViFi-B~\cite{rasheed2023vificlip} & \gc 
    \gc 67.5 & \gc 47.7 & \gc 21.8 & \gc 46.5 &
    \gc 52.6 & \gc 31.5 & \gc 31.3 \\
\midrule
\multirow{9}{*}{\begin{tabular}[c]{@{}l@{}}50\%\\Seen\\50\%\\Unseen\end{tabular}}
    & EffPrompt~\cite{ju2022effprompt} & CLIP-B~\cite{radford2021clip} &
    37.2 & 21.6 & 7.2 & 21.9 &
    32.0 & 19.3 & 19.6 \\
    & STALE~\cite{nag2022stale} & CLIP-B~\cite{radford2021clip} &
    38.3 & 21.2 & 7.0 & 22.2 &
    32.1 & 20.7 & 20.5 \\
    & UnLoc~\cite{yan2023unloc} & CLIP-B~\cite{radford2021clip} &
    - & - & - & - &
    36.9 & - & - \\
    & UnLoc~\cite{yan2023unloc} & CLIP-L~\cite{radford2021clip} &
    - & - & - & - &
    43.7 & - & - \\
    & Ju~\etal~\cite{ju2023multi} & CLIP-B~\cite{radford2021clip} &
    42.3 & 25.8 & 7.5 & 25.3 &
    34.3 & 20.8 & 21.0 \\
    \cmidrule(l){2-10}
    & \textbf{\baselineModel{}} & CLIP-B~\cite{radford2021clip} & 
    44.2 & 25.7 & 8.0 & 26.0 &
    42.1 & 25.0 & 24.8 \\ 
    & \textbf{\baselineModel{}} & ViFi-B~\cite{rasheed2023vificlip} & 
    53.4 & 31.3 & 9.8 & 31.5 &
    48.1 & 28.4 & \textbf{28.0} \\ 
    & \textbf{\ovidModel{}} & ViFi-B~\cite{rasheed2023vificlip} & 
    \textbf{56.3} & \textbf{34.4} & \textbf{11.3} & \textbf{34.0} &
    \textbf{48.4} & \textbf{28.7} & 27.9 \\ 
    & \textbf{\ovodModel{}} & ViFi-B~\cite{rasheed2023vificlip} &
    54.3 & 32.5 & 10.6 & 32.5 &
    - & - & - \\
    \cmidrule(l){2-10}
    & \gc \fullshotModel{} & \gc ViFi-B~\cite{rasheed2023vificlip} & \gc 
    \gc 68.2 & \gc 50.7 & \gc 24.6 & \gc 48.8 &
    \gc 49.4 & \gc 29.9 & \gc 29.6 \\ 
\bottomrule
\end{tabular}
}
\caption{
\textbf{Evaluation of ZS-TAL benchmark.}
The results are based on RGB only without optical flow and a single backbone model.
In each setting, the best for each metric is bolded.
Full-shot results are shown in \textcolor{gray}{gray}.
The values not provided are filled by ``-''.
Full results are presented in the supplementary.
}
\label{tab:zstal_cross_category}
\vspace{-2mm}
\end{table}

\subsection{Results of ZS-TAL Benchmark}
Tab.~\ref{tab:zstal_cross_category} presents our results on the existing ZS-TAL benchmark. For a fair comparison, we experiment with our model using CLIP backbone and without self-training (w/o ST). It shows the competitive performance, setting a strong baseline. When we integrate a video-tuned VLM, ViFi, and apply self-training, further improvements are observed.

In TH14, large improvements are achieved from self-training with ID data. For example, with a tIoU at 0.5, mAP increases as 31.3 $\rightarrow$ 34.4 and 34.3 $\rightarrow$ 37.5 on 50\%--50\% and 75\%--25\% zero-shot setting, respectively. Using OD data for self-training results in smaller improvements as the evaluation dataset is confined to the narrow sports domain.

Self-training shows only marginal improvement in ANET. This is attributed to the characteristics of ANET which mainly consists of a single long action instance, lacking the diversity of action duration. In both zero-shot settings, the mAP of ZS is close to that of FS; for instance, in the 75\%--25\% split, the mAP of ZS and FS, with a tIoU at 0.5, are 51.7 and 52.6, respectively, showing a gap of only 0.9. This indicates that ANET does not demand cross-category generalization in action localization, leading to limited gains from self-training. Based on this observation, we exclude ANET from our OV-TAL benchmark.
\section{Conclusion}
We investigated the scalability of self-training for OV-TAL. By utilizing large-scale web video data, we demonstrated large improvements in the generalizability of the action localizer, both across categories and domains. Additionally, we identified the problem of evaluation schemes in the existing ZS-TAL benchmark and proposed a more rigorous OV-TAL benchmark. Finally, we presented the results of powerful LMM Gemini 1.5 for OV-TAL and observed that it underperforms in detecting short action instances.
\vspace{-2mm}
\section*{Acknowledgement}
\vspace{-2mm}
This work was supported by the National Research Foundation of Korea (NRF) grant funded by the Korea government (MSIT) (NRF- 2022R1A2C2004509).

\appendix

\section*{Appendix}
In this section, we provide additional experimental results and details not presented in the main paper.

\section{Additional Implementation Details}
We primarily adopt the hyperparameters from ActionFormer~\cite{zhang2022actionformer} and ViFi-CLIP~\cite{rasheed2023vificlip} since our architecture is based on them.
As addressed in Sec.~4.1, the ActivityNet~\cite{caba2015activitynet} dataset differs significantly from the THUMOS14~\cite{idrees2017thumos14} and FineAction~\cite{liu2022fineaction} datasets.
It consists of only 1 -- 2 long action instances, limiting the evaluation of the capability in action localization.
Accordingly, we adjust hyperparameters for the ActivityNet dataset, following the TAL literature.

\subsection{Video Feature Extraction Details}
Here, we detail the video feature extraction process using VLMs.
As stated in Sec.~4.3, we extract the video snippet features ($\fVideo$) using the conventional sliding window manner with a window size of 16 frames and a stride size of 4 frames, after interpolating videos into 30 fps.
For ActivityNet, we interpolate $\fVideo$ to a fixed length following the widely used convention~\cite{lin2019bmn, xu2020gtad, lin2021afsd, zhang2022actionformer}.
Specifically, each video is interpolated to a length of 192 feature vectors, as employed in ActionFormer~\cite{zhang2022actionformer}.
For THUMOS14 and FineAction, we retain the video snippet features in their original length.
When we conduct experiments with ViCLIP~\cite{wang2023internvid} VLM model, we interpolate its learned temporal embedding from 8 to 16 to use the same window size for video feature extraction.
Note that such details are often absent in existing OV-TAL methods~\cite{ju2022effprompt, nag2022stale}, making it challenging to reproduce their results.
We open-source the extracted features to promote further development in the OV-TAL community.

\subsection{STOV-TAL Inference Details}
Compared to ActionFormer~\cite{zhang2022actionformer}, we use the class-agnostic action localizer and assign the action classes from the VLM.
As a result of this change, we delay the Soft-NMS operation until after the action classes are assigned, rather than immediately following the output of the action localizer.
On the other hand, when we compute pseudo labels on unlabeled videos, we directly perform Soft-NMS on the class-agnostic action instances, based on its actionness score ($s_a$).
We employ the same Soft-NMS configurations for both cases, with slight variations based on the dataset.
We choose the top 100 scoring action instances for ActivityNet and 200 for THUMOS14 and FineAction, applying a minimum confidence score threshold of 0.001 and a tIoU threshold of 0.1.

\subsection{Gemini Inference Details}
\begin{figure*}[t!]
\centering
\includegraphics[width=1.0\linewidth]{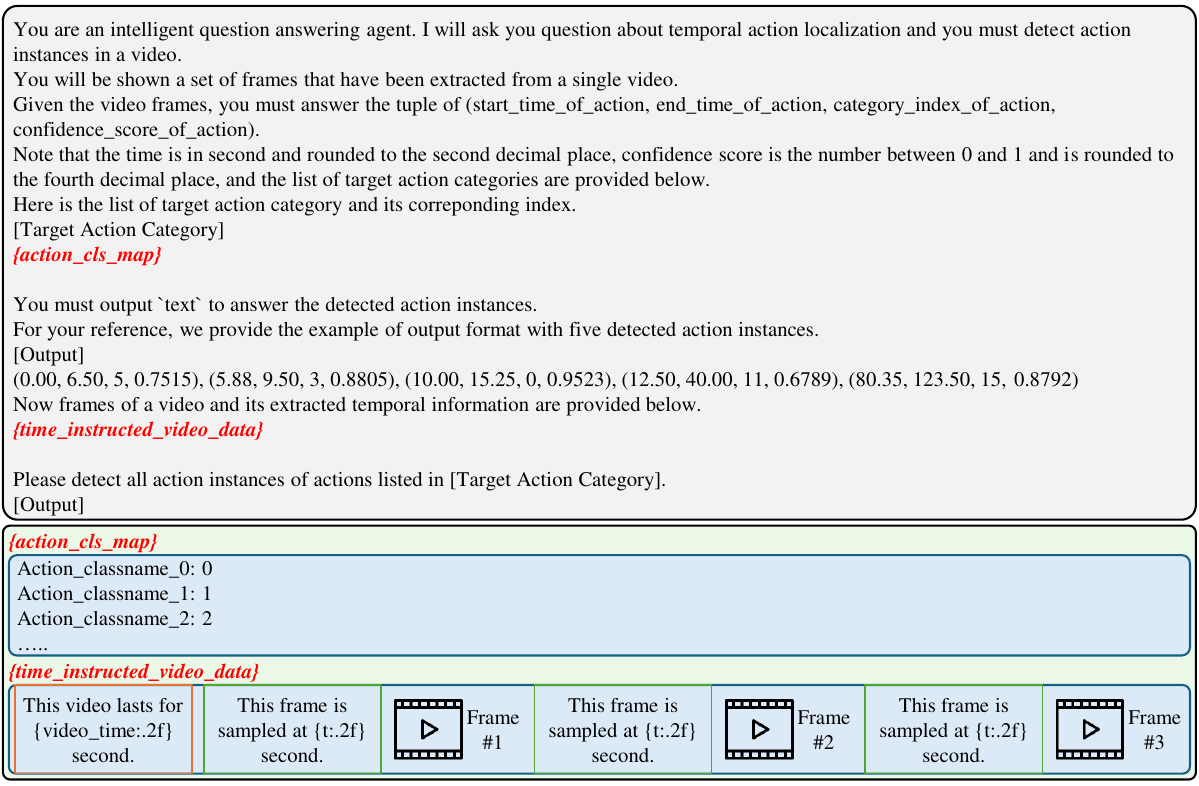}
\vspace{-4mm}
\caption{
\textbf{Gemini instruction template for TAL.}
}
\vspace{-3mm}
\label{fig:gemini_tmpl}
\end{figure*}

We provide the instruction template used for Gemini to perform the TAL task in Fig.~\ref{fig:gemini_tmpl}.
To incorporate temporal information effectively, we adopt an interleaved format, as shown in Fig.~\ref{fig:gemini_tmpl} \{time\_instructed\_video\_data\}, where RGB frame data alternates with its corresponding temporal information data throughout the entire video sequence.
This ensures that both the visual content and temporal details are presented simultaneously.
We also experimented with another format, which provides temporal information after all RGB frame data, following this structure: ``These frames are located at \{frame\_time\_list\}.''
However, Gemini was unable to effectively perform TAL with this instruction format.
The maximum length of the Gemini output is set to 4096.
For parsing Gemini's response to TAL format, we employ the regular expression of \((\mathbb{R}, \mathbb{R}, \mathbb{Z}, \mathbb{R})\), where $\mathbb{R}$ represents real numbers and $\mathbb{Z}$ represents integers.
Additionally, we filter out results where the class index falls outside the valid range, which is between 0 and $|\mathcal{C}|-1$.

Recently, Wake et al.~\cite{wake2024gpt4tal} introduced the T-PIVOT method for TAL using GPT-4o. Due to the limited context length of GPT-4o, which cannot accommodate the densely sampled frames of long videos, T-PIVOT progressively narrows the search window over time. In contrast, our Gemini-based method can detect all action instances for the target categories in a single iteration, avoiding the need for multiple API calls.

\subsection{Selection of Previous Methods for OV-TAL}
Including the results of existing OV-TAL methods~\cite{ju2022effprompt, nag2022stale, phan2024zeetad} in our proposed OV-TAL benchmarks would be a good practice to ensure a sufficient comparison.
However, due to the difficulty of reproducing their results\footnote{In \href{https://github.com/sauradip/STALE/issues/8}{Issue 1} and \href{https://github.com/sauradip/STALE/issues/23\#issuecomment-1704056365}{Issue 2}, the author mentioned that the method exploits UNet~\cite{wang2017unet} as the external action classifier which is trained on all action classes in the dataset.}, we instead choose OpenTAL~\cite{bao2022opental}, which focuses on localizing actions unseen during training through uncertainty modeling.
We train it on the base categories defined in our OV-TAL benchmark and use it as the class-agnostic action localizer in our decoupled architecture.
Although OpenTAL utilizes the I3D~\cite{carreira2017i3d} backbone for action localization, we assign action classes to output action instances using ViFi-CLIP~\cite{rasheed2023vificlip}, as same as our model.
Thus, comparing its mAPs with ours solely evaluates the capability of action localization.

In terms of EffPrompt~\cite{ju2022effprompt}, the details about the action localization model is not enough to reproduce by ourselves. 
As it also adopted decoupled architecture as ours, we use our action localizer, but replace the action classifier with the prompt-tuned VLM.
Most of the training details of prompt tuning are borrowed from EffPrompt~\cite{ju2022effprompt}, but we empirically find the training iterations for training it on THOMOS14~\cite{idrees2017thumos14} and FineAction~\cite{liu2022fineaction}.

\subsection{Training Details}
AdamW~\cite{loshchilov2018adamw} is chosen as the optimizer, coupled with a scheduler that linearly warms up the learning rate (lr) to its maximum value and decays to the minimum value ($1e^{-8}$) following a cosine function.
Tab.~\ref{tab:training_details} presents the hyperparameter values for each dataset.
For the OD-ST experiments on THUMOS14, we use the subset of 100k videos.
We empirically found the threshold values for obtaining pseudo labels.
0.2, 0.05, and 0.4 are used for ActivityNet, THUMOS14, and FineAction, respectively.

\begin{table*}[t!]
\centering
\resizebox{0.70\linewidth}{!}{
\begin{tabular}{@{} l l| cccc @{}}
\toprule
    Dataset & Stage & max lr & warm-up epoch & main epoch & batch size
    \\
\midrule
    \multirow{2}{*}{ActivityNet} 
    & First & $1e^{-5}$ & 5 & 10 & 16 \\
    & Second & $1e^{-5}$ & 5 & 5 & 16 \\
    \midrule
    \multirow{2}{*}{THUMOS14} 
    & First & $1e^{-4}$ & 5 & 30 & 2 \\
    & Second (ID-ST) & $1e^{-4}$ & 5 & 10 & 2 \\
    \midrule
    \multirow{2}{*}{FineAction} 
    & First & $1e^{-5}$ & 5 & 10 & 4 \\
    & Second (ID-ST) & $1e^{-5}$ & 5 & 5 & 4 \\
    \midrule
    THUMOS14 & \multirow{2}{*}{Second (OD-ST)} & \multirow{2}{*}{$1e^{-5}$} & \multirow{2}{*}{2} & \multirow{2}{*}{2} & \multirow{2}{*}{4} \\
    FineAction & & & & & \\
\bottomrule    
\end{tabular}
}
\vspace{-2mm}
\caption{
\textbf{Training hyperparameters values.} 
}
\label{tab:training_details}
\vspace{-2mm}
\end{table*}


\section{Additional Experimental Results}

\begin{table}[t!]
\aboverulesep=0ex
\belowrulesep=0ex
\setlength{\tabcolsep}{3pt} 
\centering
\resizebox{1.0\linewidth}{!}{
\begin{tabular}{l|l | ccc | cc }
    \toprule
    \multirow{3}{*}{Methods} & \multirow{3}{*}{Backbone} & \multicolumn{5}{c}{ActivityNet} 
    \\
    \cmidrule{3-7} & & \multicolumn{3}{c|}{Generalized} & \multicolumn{2}{c}{Constrained} 
    \\
    & & mAP$_{A}^{50}$ & mAP$_{B}^{50}$ & mAP$_{N}^{50}$ & mAP$_{B}^{50}$ & mAP$_{N}^{50}$
    \\
\midrule
\midrule
    OpenTAL$^\dagger$~\cite{bao2022opental} & I3D~\cite{carreira2017i3d} & 32.7 & 36.4 & 29.7 & 39.1 & 34.1 
    \\
    \textbf{\baselineModel{}} & ViFi-B~\cite{rasheed2023vificlip} & 42.9 & 47.5 & 39.1 & 51.3 & 44.1
    \\
    \textbf{\ovidModel{}} & ViFi-B~\cite{rasheed2023vificlip} & 43.1 & 47.3 & 39.7 & 51.1 & 44.6 
    \\ 
\midrule
    \gc \fullshotModel{} & \gc ViFi-B~\cite{rasheed2023vificlip} & \gc 43.7 & \gc 47.9 & \gc 40.3 & \gc 52.0 & \gc 45.3 
    \\
\bottomrule
\end{tabular}
}
\vspace{-2mm}
\caption{
\textbf{Evaluation of cross-category OV-TAL benchmark}.
For reference of upper-bound, full-shot results are shown in \textcolor{gray}{gray}.
$^\dagger$ is our reproduced result.
}
\label{tab:ovtal_cross_cateogry_anet}
\vspace{-2mm}
\end{table}

\begin{table*}[t!]
\aboverulesep=0ex
\belowrulesep=0ex
\centering
\resizebox{1.0\linewidth}{!}{
\begin{tabular}{ l|l|l| ccccc|c | ccc|c }
\toprule
  \multirow{2}{*}{\begin{tabular}{@{}l@{}} Evaluation \\ Setting \end{tabular}} &
  \multirow{2}{*}{Methods} &
  \multirow{2}{*}{Backbone} &
  \multicolumn{6}{c|}{THUMOS14~\cite{idrees2017thumos14}} & 
  \multicolumn{4}{c}{ActivityNet v1.3~\cite{caba2015activitynet}} \\
  \cmidrule{4-9} \cmidrule{10-13} 
  & & & 0.3 & 0.4 & 0.5 & 0.6 & 0.7 & Avg. & 0.5 & 0.75 & 0.95 & Avg. \\
\midrule
\midrule
\multirow{7}{*}{\begin{tabular}[c]{@{}l@{}}Full-shot\\ 100\% Seen\\ 0\% Unseen \end{tabular}}
    & ActionFormer$^\dagger$~\cite{zhang2022actionformer} & ViFi-CLIP-B~\cite{rasheed2023vificlip} & 72.8 & 67.4 & 57.3 & 45.2 & 29.7 & 54.5 & 49.5 & 31.2 & 4.3 & 30.9 \\
    & EffPrompt~\cite{ju2022effprompt} & CLIP-B~\cite{radford2021clip} & 50.8 & 44.1 & 35.8 & 25.7 & 15.7 & 34.5 & 44.0 & 27.0 & 5.1 & 27.3 \\
    & STALE~\cite{nag2022stale} & CLIP-B~\cite{radford2021clip} & 60.6 & 53.2 & 44.6 & 36.8 & 26.7 & 44.4 & 54.3 & 34.0 & 7.7 & 34.3 \\
    & UnLoc~\cite{yan2023unloc} & CLIP-B~\cite{radford2021clip} & - & - & - & - & - & - & 54.6 & - & - & - \\
    & UnLoc~\cite{yan2023unloc} & CLIP-L~\cite{radford2021clip}& - & - & - & - & - & - & 59.3 & - & - & - \\
    \cmidrule(lr){2-13}
    & \textbf{\fullshotModel{}} & CLIP-B~\cite{radford2021clip}& 47.5 & 41.7 & 33.0 & 24.7 & 15.4 & 32.5 & 37.5 & 23.1 & 1.8 & 22.7 \\
    & \textbf{\fullshotModel{}} & ViFi-CLIP-B~\cite{rasheed2023vificlip} & 65.3 & 60.6 & 50.2 & 39.0 & 26.0 & 48.2 & 43.7 & 26.8 & 2.0 & 26.4 \\
\midrule
\multirow{9}{*}{\begin{tabular}[c]{@{}l@{}}Zero-shot\\ 75\% Seen\\ 25\% Unseen \end{tabular}}
    & EffPrompt~\cite{ju2022effprompt} & CLIP-B~\cite{radford2021clip} & 39.7 & 31.6 & 23.0 & 14.9 & 7.5 & 23.3 & 37.6 & 22.9 & 3.8 & 23.1 \\
    & STALE~\cite{nag2022stale} & CLIP-B~\cite{radford2021clip} & 40.5 & 32.3 & 23.5 & 15.3 & 7.6 & 23.8 & 38.2 & 25.2 & \textbf{6.0}& 24.9 \\
    & UnLoc~\cite{yan2023unloc} & CLIP-B~\cite{radford2021clip} & - & - & - & - & - & - & 40.2 & - & - & - \\
    & UnLoc~\cite{yan2023unloc} & CLIP-L~\cite{radford2021clip}& - & - & - & - & - & - & 48.8 & - & - & - \\
    & Ju~\etal~\cite{ju2023multi} & CLIP-B~\cite{radford2021clip} & 46.3 & 39.0 & 29.5 & 18.3 & 8.7 & 28.4 & 42.0 & 25.8 & 3.2 & 25.9 \\
    \cmidrule(lr){2-13}
    & \textbf{\baselineModel{}} & CLIP-B~\cite{radford2021clip} & 
    47.8 & 39.1 & 28.4 & 17.6 & 9.1 & 28.4 & 
    47.0 & 28.1 & 1.6 & 27.9 \\   
    & \textbf{\baselineModel{}} & ViFi-CLIP-B~\cite{rasheed2023vificlip} & 
    56.7 & 47.2 & 34.3 & 22.8 & 11.3 & 34.5 &
    51.7 & \textbf{30.9} & 1.8 & \textbf{30.5} \\
    & \textbf{\ovidModel{}} & ViFi-CLIP-B~\cite{rasheed2023vificlip} & 
    \textbf{59.5} & \textbf{50.2} & \textbf{37.5} & \textbf{24.6} & \textbf{12.5} & \textbf{36.9} &
    \textbf{52.0} & 30.6 & 1.2 & 30.1 
    \\ 
    & \textbf{\ovodModel{}} & ViFi-CLIP-B~\cite{rasheed2023vificlip} &58.2 & 48.2 & 35.1 & 23.0 & 11.8 & 35.2 & - & - & - & -
    \\
    \cmidrule(l){2-13}
    & \gc \fullshotModel{} & \gc ViFi-CLIP-B~\cite{rasheed2023vificlip} & \gc 
    \gc 67.5 & \gc 60.8 & \gc 47.7 & \gc 34.8 & \gc 21.8 & \gc 46.5 &
    \gc 52.6 & \gc 31.5 & \gc 2.3 & \gc 31.3 \\
\midrule
\multirow{9}{*}{\begin{tabular}[c]{@{}l@{}}Zero-shot\\ 50\% Seen\\ 50\% Unseen\end{tabular}}
    & EffPrompt~\cite{ju2022effprompt} & CLIP-B~\cite{radford2021clip} & 37.2 & 29.6 & 21.6 & 14.0 & 7.2 & 21.9 & 32.0 & 19.3 & 2.9 & 19.6 \\
    & STALE~\cite{nag2022stale} & CLIP-B~\cite{radford2021clip} & 38.3 & 30.7 & 21.2 & 13.8 & 7.0 & 22.2 & 32.1 & 20.7 & \textbf{5.9} & 20.5 \\
    & UnLoc~\cite{yan2023unloc} & CLIP-B~\cite{radford2021clip} & - & - & - & - & - & - & 36.9 & - & - & - \\
    & UnLoc~\cite{yan2023unloc} & CLIP-L~\cite{radford2021clip}& - & - & - & - & - & - & 43.7 & - & - & - \\
    & Ju~\etal~\cite{ju2023multi} & CLIP-B~\cite{radford2021clip} & 42.3 & 34.7 & 25.8 & 16.2 & 7.5 & 25.3 & 34.3 & 20.8 & 3.0 & 21.0 \\
    \cmidrule(l){2-13}
    & \textbf{\baselineModel{}} & CLIP-B~\cite{radford2021clip} & 
    44.2 & 35.7 & 25.7 & 16.5 & 8.0 & 26.0 &
    42.1 & 25.0 & 1.3 & 24.8 \\ 
    & \textbf{\baselineModel{}} & ViFi-CLIP-B~\cite{rasheed2023vificlip} & 
    53.4 & 43.1 & 31.3 & 19.7 & 9.8 & 31.5 &
    48.1 & 28.4 & 1.3 & \textbf{28.0} \\ 
    & \textbf{\ovidModel{}} & ViFi-CLIP-B~\cite{rasheed2023vificlip} & 
    \textbf{56.3} & \textbf{46.1} & \textbf{34.4} & \textbf{21.9} & \textbf{11.3} & \textbf{34.0} &
    \textbf{48.4} & \textbf{28.7} & 0.8 & 27.9 
    \\
    & \textbf{\ovodModel{}} & ViFi-CLIP-B~\cite{rasheed2023vificlip} & 54.3 & 43.7 & 32.5 & 21.4 & 10.6 & 32.5 & - & - & - & -
    \\
    \cmidrule(l){2-13}
    & \gc \fullshotModel{} & \gc ViFi-CLIP-B~\cite{rasheed2023vificlip} & \gc 
    \gc 68.2 & \gc 62.5 & \gc 50.7 & \gc 38.2 & \gc 24.6 & \gc 48.8 &
    \gc 49.4 & \gc 29.9 & \gc 2.2 & \gc 29.6 \\ 
\bottomrule
\end{tabular}
}
\vspace{-2mm}
\caption{
\textbf{Evaluation of ZS-TAL benchmark.}
The results are based on RGB only without optical flow.
In each setting, the best for each metric is bolded.
Full-shot results are shown in \textcolor{gray}{gray} for reference of upper-bound.
$^\dagger$ indicates our reproduced results.
The values not provided are filled by ``-''.
}
\label{tab:zstal_cross_category_full}
\vspace{-2mm}
\end{table*}

\subsection{ActivityNet Results for OV-TAL Benchmark}
The main paper presents the cross-category OV-TAL benchmark results of the THUMOS14~\cite{idrees2017thumos14} and FineAction~\cite{liu2022fineaction} datasets in Tab.~4 (Main).
Here, we show the results of the ActivityNet v1.3~\cite{caba2015activitynet} in Tab.~\ref{tab:ovtal_cross_cateogry_anet}.
As discussed in Sec.~4.6, ActivityNet v1.3 is not a proper dataset for evaluating the generalization capability in action localization, supported by the zero-shot performance (w/o ST) of action localization on par with that of full-shot in Tab.~6 (Main).
In the proposed OV-TAL benchmark, we observe a similar trend.
For instance, in the constrained setting, mAP$^{50}_{N}$ of w/o ST and FS are 44.1 and 45.3, respectively.
Based on these results, we decided not to include the ActivityNet dataset in the OV-TAL benchmark since there is only a small room for improvement in cross-category generalization ability.

\subsection{ZS-TAL Benchmark Full Results}
Due to space constraints, we present only partial results of the ZS-TAL benchmark in Tab.~6 (Main). In Tab.~\ref{tab:zstal_cross_category_full}, we provide the complete results, which complement the tIoU values of 0.4 and 0.6 for TH14, and 0.95 for ANET. These results exhibit a similar trend to those presented in the main paper. In the case of full-shot results (100\% Seen 0\% Unseen), other methods achieve higher mAP, which is attributed to the use of fine-tuned classifiers. These methods fine-tune the action classifiers on the target action categories, resulting in identical action categories during training and testing. In contrast, we keep freeze and do not fine-tune the VLM for the target actions, and ours perform better for zero-shot settings. Therefore, the 100\% Seen 0\% Unseen results do not reflect the generalization ability of action localizers.

{\small
\bibliographystyle{ieee_fullname}
\bibliography{main}

\begin{thebibliography}{10}\itemsep=-1pt

\bibitem{bao2022opental}
Wentao Bao, Qi Yu, and Yu Kong.
\newblock Opental: Towards open set temporal action localization.
\newblock In {\em CVPR}, pages 2979--2989, 2022.

\bibitem{bird2009nltk}
Steven Bird, Ewan Klein, and Edward Loper.
\newblock {\em Natural language processing with Python: analyzing text with the natural language toolkit}.
\newblock " O'Reilly Media, Inc.", 2009.

\bibitem{bodla2017softnms}
Navaneeth Bodla, Bharat Singh, Rama Chellappa, and Larry~S Davis.
\newblock Soft-nms--improving object detection with one line of code.
\newblock In {\em ICCV}, pages 5561--5569, 2017.

\bibitem{brown2020gpt3}
Tom Brown, Benjamin Mann, Nick Ryder, Melanie Subbiah, Jared~D Kaplan, Prafulla Dhariwal, Arvind Neelakantan, Pranav Shyam, Girish Sastry, Amanda Askell, et~al.
\newblock Language models are few-shot learners.
\newblock {\em NeurIPS}, 33:1877--1901, 2020.

\bibitem{caba2015activitynet}
Fabian Caba~Heilbron, Victor Escorcia, Bernard Ghanem, and Juan Carlos~Niebles.
\newblock Activitynet: A large-scale video benchmark for human activity understanding.
\newblock In {\em CVPR}, pages 961--970, 2015.

\bibitem{carreira2018k600}
Joao Carreira, Eric Noland, Andras Banki-Horvath, Chloe Hillier, and Andrew Zisserman.
\newblock A short note about kinetics-600.
\newblock {\em arXiv preprint arXiv:1808.01340}, 2018.

\bibitem{carreira2017i3d}
Joao Carreira and Andrew Zisserman.
\newblock Quo vadis, action recognition? a new model and the kinetics dataset.
\newblock In {\em CVPR}, pages 6299--6308, 2017.

\bibitem{cherti2023openclip}
Mehdi Cherti, Romain Beaumont, Ross Wightman, Mitchell Wortsman, Gabriel Ilharco, Cade Gordon, Christoph Schuhmann, Ludwig Schmidt, and Jenia Jitsev.
\newblock Reproducible scaling laws for contrastive language-image learning.
\newblock In {\em CVPR}, pages 2818--2829, 2023.

\bibitem{deng2009imagenet}
Jia Deng, Wei Dong, Richard Socher, Li-Jia Li, Kai Li, and Li Fei-Fei.
\newblock Imagenet: A large-scale hierarchical image database.
\newblock In {\em CVPR}, pages 248--255. IEEE, 2009.

\bibitem{devlin2019bert}
Jacob Devlin, Ming-Wei Chang, Kenton Lee, and Kristina Toutanova.
\newblock Bert: Pre-training of deep bidirectional transformers for language understanding.
\newblock In {\em NACCL}, pages 4171--4186, 2019.

\bibitem{dosovitskiy2021vit}
Alexey Dosovitskiy, Lucas Beyer, Alexander Kolesnikov, Dirk Weissenborn, Xiaohua Zhai, Thomas Unterthiner, Mostafa Dehghani, Matthias Minderer, Georg Heigold, Sylvain Gelly, Jakob Uszkoreit, and Neil Houlsby.
\newblock An image is worth 16x16 words: Transformers for image recognition at scale.
\newblock In {\em ICLR}, 2021.

\bibitem{he2017mask-rcnn}
Kaiming He, Georgia Gkioxari, Piotr Doll{\'a}r, and Ross Girshick.
\newblock Mask r-cnn.
\newblock In {\em ICCV}, pages 2961--2969, 2017.

\bibitem{idrees2017thumos14}
Haroon Idrees, Amir~R Zamir, Yu-Gang Jiang, Alex Gorban, Ivan Laptev, Rahul Sukthankar, and Mubarak Shah.
\newblock The thumos challenge on action recognition for videos “in the wild”.
\newblock {\em CVIU}, 155:1--23, 2017.

\bibitem{jia2021align}
Chao Jia, Yinfei Yang, Ye Xia, Yi-Ting Chen, Zarana Parekh, Hieu Pham, Quoc Le, Yun-Hsuan Sung, Zhen Li, and Tom Duerig.
\newblock Scaling up visual and vision-language representation learning with noisy text supervision.
\newblock In {\em ICML}, pages 4904--4916. PMLR, 2021.

\bibitem{ju2022effprompt}
Chen Ju, Tengda Han, Kunhao Zheng, Ya Zhang, and Weidi Xie.
\newblock Prompting visual-language models for efficient video understanding.
\newblock In {\em ECCV}, pages 105--124. Springer, 2022.

\bibitem{ju2023multi}
Chen Ju, Zeqian Li, Peisen Zhao, Ya Zhang, Xiaopeng Zhang, Qi Tian, Yanfeng Wang, and Weidi Xie.
\newblock Multi-modal prompting for low-shot temporal action localization.
\newblock {\em arXiv preprint arXiv:2303.11732}, 2023.

\bibitem{li2022rethinking}
Hengduo Li, Zuxuan Wu, Abhinav Shrivastava, and Larry~S Davis.
\newblock Rethinking pseudo labels for semi-supervised object detection.
\newblock In {\em AAAI}, volume~36, pages 1314--1322, 2022.

\bibitem{li2022blip}
Junnan Li, Dongxu Li, Caiming Xiong, and Steven Hoi.
\newblock Blip: Bootstrapping language-image pre-training for unified vision-language understanding and generation.
\newblock In {\em ICML}, pages 12888--12900. PMLR, 2022.

\bibitem{li2024detal}
Zhiheng Li, Yujie Zhong, Ran Song, Tianjiao Li, Lin Ma, and Wei Zhang.
\newblock Detal: Open-vocabulary temporal action localization with decoupled networks.
\newblock {\em TPAMI}, 2024.

\bibitem{lin2021afsd}
Chuming Lin, Chengming Xu, Donghao Luo, Yabiao Wang, Ying Tai, Chengjie Wang, Jilin Li, Feiyue Huang, and Yanwei Fu.
\newblock Learning salient boundary feature for anchor-free temporal action localization.
\newblock In {\em CVPR}, pages 3320--3329, 2021.

\bibitem{lin2019bmn}
Tianwei Lin, Xiao Liu, Xin Li, Errui Ding, and Shilei Wen.
\newblock Bmn: Boundary-matching network for temporal action proposal generation.
\newblock In {\em ICCV}, pages 3889--3898, 2019.

\bibitem{lin2017focal}
Tsung-Yi Lin, Priya Goyal, Ross Girshick, Kaiming He, and Piotr Doll{\'a}r.
\newblock Focal loss for dense object detection.
\newblock In {\em ICCV}, pages 2980--2988, 2017.

\bibitem{lin2023MAXI}
Wei Lin, Leonid Karlinsky, Nina Shvetsova, Horst Possegger, Mateusz Kozinski, Rameswar Panda, Rogerio Feris, Hilde Kuehne, and Horst Bischof.
\newblock Match, expand and improve: Unsupervised finetuning for zero-shot action recognition with language knowledge.
\newblock In {\em ICCV}, pages 2851--2862, 2023.

\bibitem{liu2022fineaction}
Yi Liu, Limin Wang, Yali Wang, Xiao Ma, and Yu Qiao.
\newblock Fineaction: A fine-grained video dataset for temporal action localization.
\newblock {\em IEEE TIP}, 31:6937--6950, 2022.

\bibitem{long2020ahernet}
Fuchen Long, Ting Yao, Zhaofan Qiu, Xinmei Tian, Jiebo Luo, and Tao Mei.
\newblock Learning to localize actions from moments.
\newblock In {\em ECCV}, pages 137--154. Springer, 2020.

\bibitem{loshchilov2018adamw}
Ilya Loshchilov and Frank Hutter.
\newblock Decoupled weight decay regularization.
\newblock In {\em ICLR}, 2019.

\bibitem{mikolov2013word2vec}
Tomas Mikolov, Ilya Sutskever, Kai Chen, Greg~S Corrado, and Jeff Dean.
\newblock Distributed representations of words and phrases and their compositionality.
\newblock {\em NeurIPS}, 26, 2013.

\bibitem{nag2023tranzad}
Sayak Nag, Orpaz Goldstein, and Amit~K Roy-Chowdhury.
\newblock Semantics guided contrastive learning of transformers for zero-shot temporal activity detection.
\newblock In {\em WACV}, pages 6243--6253, 2023.

\bibitem{nag2022stale}
Sauradip Nag, Xiatian Zhu, Yi-Zhe Song, and Tao Xiang.
\newblock Zero-shot temporal action detection via vision-language prompting.
\newblock In {\em ECCV}, pages 681--697. Springer, 2022.

\bibitem{ni2022xclip}
Bolin Ni, Houwen Peng, Minghao Chen, Songyang Zhang, Gaofeng Meng, Jianlong Fu, Shiming Xiang, and Haibin Ling.
\newblock Expanding language-image pretrained models for general video recognition.
\newblock In {\em ECCV}. Springer, 2022.

\bibitem{park2021tcd}
Jaeyoo Park, Minsoo Kang, and Bohyung Han.
\newblock Class-incremental learning for action recognition in videos.
\newblock In {\em ICCV}, pages 13698--13707, 2021.

\bibitem{pham2023basic}
Hieu Pham, Zihang Dai, Golnaz Ghiasi, Kenji Kawaguchi, Hanxiao Liu, Adams~Wei Yu, Jiahui Yu, Yi-Ting Chen, Minh-Thang Luong, Yonghui Wu, Mingxing Tan, and Quoc~V. Le.
\newblock Combined scaling for zero-shot transfer learning.
\newblock {\em Neurocomputing}, 555:126658, 2023.

\bibitem{phan2024zeetad}
Thinh Phan, Khoa Vo, Duy Le, Gianfranco Doretto, Donald Adjeroh, and Ngan Le.
\newblock Zeetad: Adapting pretrained vision-language model for zero-shot end-to-end temporal action detection.
\newblock In {\em WACV}, pages 7046--7055, 2024.

\bibitem{radford2021clip}
Alec Radford, Jong~Wook Kim, Chris Hallacy, Aditya Ramesh, Gabriel Goh, Sandhini Agarwal, Girish Sastry, Amanda Askell, Pamela Mishkin, Jack Clark, et~al.
\newblock Learning transferable visual models from natural language supervision.
\newblock In {\em ICML}, volume 139, pages 8748--8763. PMLR, 2021.

\bibitem{rasheed2023vificlip}
Hanoona Rasheed, Muhammad~Uzair Khattak, Muhammad Maaz, Salman Khan, and Fahad~Shahbaz Khan.
\newblock Fine-tuned clip models are efficient video learners.
\newblock In {\em CVPR}, pages 6545--6554, 2023.

\bibitem{rathod2022ovtad}
Vivek Rathod, Bryan Seybold, Sudheendra Vijayanarasimhan, Austin Myers, Xiuye Gu, Vighnesh Birodkar, and David~A Ross.
\newblock Open-vocabulary temporal action detection with off-the-shelf image-text features.
\newblock {\em arXiv preprint arXiv:2212.10596}, 2022.

\bibitem{reid2024gemini}
Machel Reid, Nikolay Savinov, Denis Teplyashin, Dmitry Lepikhin, Timothy Lillicrap, Jean-baptiste Alayrac, Radu Soricut, Angeliki Lazaridou, Orhan Firat, Julian Schrittwieser, et~al.
\newblock Gemini 1.5: Unlocking multimodal understanding across millions of tokens of context.
\newblock {\em arXiv preprint arXiv:2403.05530}, 2024.

\bibitem{ren2024timechat}
Shuhuai Ren, Linli Yao, Shicheng Li, Xu Sun, and Lu Hou.
\newblock Timechat: A time-sensitive multimodal large language model for long video understanding.
\newblock In {\em CVPR}, pages 14313--14323, 2024.

\bibitem{sennrich2016bpe}
Rico Sennrich, Barry Haddow, and Alexandra Birch.
\newblock Neural machine translation of rare words with subword units.
\newblock In {\em ACL}, volume~1, pages 1715--1725, 2016.

\bibitem{tarvainen2017mean-teacher}
Antti Tarvainen and Harri Valpola.
\newblock Mean teachers are better role models: Weight-averaged consistency targets improve semi-supervised deep learning results.
\newblock {\em NeurIPS}, 30, 2017.

\bibitem{wake2024gpt4tal}
Naoki Wake, Atsushi Kanehira, Kazuhiro Sasabuchi, Jun Takamatsu, and Katsushi Ikeuchi.
\newblock Open-vocabulary temporal action localization using vlms.
\newblock {\em arXiv preprint arXiv:2408.17422}, 2024.

\bibitem{wang2017unet}
Limin Wang, Yuanjun Xiong, Dahua Lin, and Luc Van~Gool.
\newblock Untrimmednets for weakly supervised action recognition and detection.
\newblock In {\em CVPR}, pages 4325--4334, 2017.

\bibitem{wang2023freematch}
Yidong Wang, Hao Chen, Qiang Heng, Wenxin Hou, Yue Fan, Zhen Wu, Jindong Wang, Marios Savvides, Takahiro Shinozaki, Bhiksha Raj, Bernt Schiele, and Xing Xie.
\newblock Freematch: Self-adaptive thresholding for semi-supervised learning.
\newblock In {\em ICLR}, 2023.

\bibitem{wang2023internvid}
Yi Wang, Yinan He, Yizhuo Li, Kunchang Li, Jiashuo Yu, Xin Ma, Xinyuan Chen, Yaohui Wang, Ping Luo, Ziwei Liu, et~al.
\newblock Internvid: A large-scale video-text dataset for multimodal understanding and generation.
\newblock {\em arXiv preprint arXiv:2307.06942}, 2023.

\bibitem{weng2023openvclip}
Zejia Weng, Xitong Yang, Ang Li, Zuxuan Wu, and Yu-Gang Jiang.
\newblock Open-{VCLIP}: Transforming {CLIP} to an open-vocabulary video model via interpolated weight optimization.
\newblock In {\em ICML}, volume 202, pages 36978--36989. PMLR, 2023.

\bibitem{wu2024ovsurvey}
Jianzong Wu, Xiangtai Li, Shilin Xu, Haobo Yuan, Henghui Ding, Yibo Yang, Xia Li, Jiangning Zhang, Yunhai Tong, Xudong Jiang, et~al.
\newblock Towards open vocabulary learning: A survey.
\newblock {\em IEEE TPAMI}, 2024.

\bibitem{xia2023npm}
Kun Xia, Le Wang, Sanping Zhou, Gang Hua, and Wei Tang.
\newblock Learning from noisy pseudo labels for semi-supervised temporal action localization.
\newblock In {\em ICCV}, pages 10160--10169, October 2023.

\bibitem{xian2018zero}
Yongqin Xian, Christoph~H Lampert, Bernt Schiele, and Zeynep Akata.
\newblock Zero-shot learning—a comprehensive evaluation of the good, the bad and the ugly.
\newblock {\em IEEE TPAMI}, 41(9):2251--2265, 2018.

\bibitem{xu2020gtad}
Mengmeng Xu, Chen Zhao, David~S Rojas, Ali Thabet, and Bernard Ghanem.
\newblock G-tad: Sub-graph localization for temporal action detection.
\newblock In {\em CVPR}, pages 10156--10165, 2020.

\bibitem{yan2023unloc}
Shen Yan, Xuehan Xiong, Arsha Nagrani, Anurag Arnab, Zhonghao Wang, Weina Ge, David Ross, and Cordelia Schmid.
\newblock Unloc: A unified framework for video localization tasks.
\newblock In {\em ICCV}, pages 13623--13633, 2023.

\bibitem{zareian2021ovrcnn}
Alireza Zareian, Kevin~Dela Rosa, Derek~Hao Hu, and Shih-Fu Chang.
\newblock Open-vocabulary object detection using captions.
\newblock In {\em CVPR}, pages 14393--14402, 2021.

\bibitem{zhang2022actionformer}
Chen-Lin Zhang, Jianxin Wu, and Yin Li.
\newblock Actionformer: Localizing moments of actions with transformers.
\newblock In {\em ECCV}, pages 492--510. Springer, 2022.

\bibitem{zhang2020zstad}
Lingling Zhang, Xiaojun Chang, Jun Liu, Minnan Luo, Sen Wang, Zongyuan Ge, and Alexander Hauptmann.
\newblock Zstad: Zero-shot temporal activity detection.
\newblock In {\em CVPR}, pages 879--888, 2020.

\bibitem{zhang2023owtal}
Yaru Zhang, Xiao-Yu Zhang, and Haichao Shi.
\newblock Ow-tal: learning unknown human activities for open-world temporal action localization.
\newblock {\em Pattern Recognition}, 133:109027, 2023.

\bibitem{zheng2020diou}
Zhaohui Zheng, Ping Wang, Wei Liu, Jinze Li, Rongguang Ye, and Dongwei Ren.
\newblock Distance-iou loss: Faster and better learning for bounding box regression.
\newblock In {\em AAAI}, volume~34, pages 12993--13000, 2020.

\bibitem{zhou2024apl}
Feixiang Zhou, Bryan Williams, and Hossein Rahmani.
\newblock Towards adaptive pseudo-label learning for semi-supervised temporal action localization.
\newblock {\em arXiv preprint arXiv:2407.07673}, 2024.

\end{thebibliography}
}

\end{document}